\documentclass[runningheads]{llncs}

\usepackage[T1]{fontenc}

\usepackage{graphicx}

\usepackage{hyperref}

\usepackage{booktabs}
\usepackage[table,xcdraw]{xcolor}
\usepackage{multirow}
\usepackage{xcolor}
\usepackage{subcaption}
\usepackage{listings}
\usepackage{tikz}

\definecolor{darkgreen}{cmyk}{1, 0, 1, 0.5}

\begin{document}
\title{Knowledge-Driven Hallucination in Large Language Models: An Empirical Study on Process Modeling\thanks{The Version of Record of this contribution will be published in the proceedings of the 2nd International Workshop on Generative AI for Process Mining (GenAI4PM 2025). This preprint has not undergone peer review or any post-submission improvements or corrections.}}

\titlerunning{Knowledge-Driven Hallucination in Large Language Models}

\author{Humam Kourani\inst{1,2}\orcidID{0000-0003-2375-2152} \and
Anton Antonov\inst{1,2}\orcidID{0009-0004-1044-4884} \and
Alessandro Berti\inst{2}\orcidID{0000-0002-3279-4795} \and 
Wil M.P. van der Aalst\inst{2,1}\orcidID{0000-0002-0955-6940}
}

\authorrunning{H. Kourani et al.}

\institute{Fraunhofer Institute for Applied Information Technology FIT, Schloss Birlinghoven, 53757 Sankt Augustin, Germany
\email{\{humam.kourani,anton.antonov\}@fit.fraunhofer.de} \and
RWTH Aachen University, Ahornstraße 55, 52074 Aachen, Germany\\
\email{\{a.berti,wvdaalst\}@pads.rwth-aachen.de}
}

\maketitle

\begin{abstract}
The utility of Large Language Models (LLMs) in analytical tasks is rooted in their vast pre-trained knowledge, which allows them to interpret ambiguous inputs and infer missing information. However, this same capability introduces a critical risk of what we term \emph{knowledge-driven hallucination}: a phenomenon where the model's output contradicts explicit source evidence because it is overridden by the model's generalized internal knowledge. This paper investigates this phenomenon by evaluating LLMs on the task of \emph{automated process modeling}, where the goal is to generate a formal business process model from a given source artifact. The domain of Business Process Management (BPM) provides an ideal context for this study, as many core business processes follow standardized patterns, making it likely that LLMs possess strong pre-trained schemas for them. We conduct a controlled experiment designed to create scenarios with deliberate conflict between provided evidence and the LLM's background knowledge. We use inputs describing both standard and deliberately atypical process structures to measure the LLM's fidelity to the provided evidence. Our work provides a methodology for assessing this critical reliability issue and raises awareness of the need for rigorous validation of AI-generated artifacts in any evidence-based domain.


\end{abstract}

\section{Introduction}
The integration of Large Language Models (LLMs) and other foundation models into analytical and data-driven workflows promises to automate complex tasks and democratize access to specialized domains. A key capability driving this transformation is the models' ability to leverage vast, pre-trained knowledge to interpret ambiguous inputs, infer missing details, and generate coherent, structured outputs. This capacity for \emph{intelligent inference} is fundamental to their performance across a range of applications, from code generation to data analysis.

Reliance of an LLM on its internal knowledge base introduces a critical challenge that we term \emph{knowledge-driven hallucination}. This conflict arises when the explicit evidence provided in a user's prompt (e.g., a source document, a dataset, or a simple text) is inconsistent with the generalized patterns and ``common sense'' knowledge the model has acquired during its training. In such situations, the model faces a dilemma: should it remain faithful to the provided evidence, even if it appears anomalous or counter-intuitive, or should it ``correct'' the output based on its pre-trained understanding of what is typical or plausible?

The outcome of this conflict poses implications for the trustworthiness and reliability of AI-driven systems. An LLM that prioritizes its internal knowledge over explicit evidence may generate outputs that are dangerously misleading. The generated artifact might appear well-formed, logical, and plausible, yet fail to accurately represent the specific reality of the input data. This risk is particularly acute in specialized domains where processes, rules, or data may be intentionally unconventional and deviate from established norms.

This paper investigates the knowledge-driven hallucination of LLMs through a systematic, empirical study within the domain of \emph{Business Process Management (BPM)}. The task of \emph{process modeling} (i.e., generating a formal process model from a source artifact) is particularly well-suited for our investigation due to the nature of business processes themselves. Many core business operations, such as purchase-to-pay, order-to-cash, or incident management, follow well-established, standardized patterns across different organizations. Consequently, it is highly probable that an LLM has been exposed to extensive documentation and descriptions of these standard processes during its training, endowing it with a strong pre-trained ``schema'' of how such processes ``should'' operate. This creates a powerful and realistic dilemma for our study: what occurs when the evidence provided for a specific organization's process directly contradicts the generalized, common-sense model of that process residing within the LLM?

To isolate and quantify this conflict, we conduct a controlled experiment using a set of standard process models ($M^+$) that represent conventional process flows. For each standard model, we create two deliberately conflicting variants: a reversed model ($M^-$), where the sequence of activities is causally inverted, and a shuffled model ($M^*$), where the original activity labels are randomly permuted across the model's structure. We then task an LLM to generate process models from textual descriptions and event logs derived from these variants. By comparing the generated models against both the source evidence and the conventional standard model, we measure the degree to which the LLM adheres to the provided evidence versus reverting to its internal knowledge.

The structure of this paper is as follows. \autoref{sec:related_work} discusses related work on process modeling and LLM hallucinations. \autoref{sec:methodology} details our experimental methodology, while \autoref{sec:results} presents our findings. Finally, \autoref{sec:conclusion} concludes the paper and discusses the broader implications of our work.

\section{Related Work}
\label{sec:related_work}
This section reviews related research, focusing on LLM hallucination and process modeling techniques.

\subsection{LLM Hallucination}
The phenomenon of \textit{hallucination} in LLMs—where models generate outputs that appear plausible yet are factually incorrect—has been extensively studied in the literature~\cite{DBLP:journals/csur/JiLFYSXIBMF23}. The underlying causes of such behavior can broadly be categorized into three groups: hallucinations arising from data, from training, and from inference~\cite{DBLP:journals/tois/HuangYMZFWCPFQL25}.

LLMs are trained on two main types of data: pre-training data, which imparts general and factual knowledge~\cite{DBLP:conf/nips/ZhouLX0SMMEYYZG23}, and alignment data, which instructs models to follow human preferences and respond to user intent~\cite{DBLP:journals/corr/abs-2307-12966}. However, the pre-training corpus is inherently limited and often biased toward general knowledge~\cite{DBLP:journals/corr/abs-2101-00027}, restricting the model’s ability to generalize to domain-specific queries~\cite{DBLP:journals/corr/abs-2305-18703}.

The training process itself imposes further constraints. Pre-training limits the effective context length a model can utilize during inference~\cite{roberts2024extending,DBLP:conf/acl/ChenLHZSLLY24}, leading to incomplete conditioning on the user’s input. Moreover, fine-tuning with human feedback may encourage the model to favor responses that align with user preferences, even when they are not strictly truthful~\cite{DBLP:conf/iclr/HoskingBB24,DBLP:conf/iclr/SharmaTKDABDHJK24}.

At inference time, LLMs generate output in an autoregressive manner, predicting the next token based on previously generated tokens and their internalized knowledge~\cite{DBLP:conf/nips/BrownMRSKDNSSAA20}. As responses become longer, models are increasingly prone to forgetting earlier parts of the prompt~\cite{DBLP:journals/air/BlancoJusticiaJMSDCT25}, which can degrade the coherence and factuality of their outputs. Apart from that, flawed reasoning may also introduce hallucinations. For instance, Berglund et al.~\cite{DBLP:conf/iclr/BerglundTKBSKE24} identify the ``Reversal Curse'', where a model that correctly answers ``A is B'' may fail when asked to infer ``B is A''.

\subsection{Process Modeling}
Business process modeling is the structured representation of the tasks, decisions, and flow within a business process \cite{DBLP:books/sp/DumasRMR18}. Organizations often rely on process models to document their workflows, and the creation of these models typically involves collaboration between business analysts and domain experts to ensure clarity and accuracy \cite{DBLP:conf/caise/ForsterPW13}. 

Traditionally, creating business process models required substantial manual effort and expertise in complex modeling languages. To automate this, early approaches primarily relied on traditional Natural Language Processing (NLP) and rule-based techniques \cite{DBLP:journals/eis/WoenselM24}. These methods exploited dependency parsing, part-of-speech tagging, and semantic role labeling to identify process elements from unstructured text \cite{DBLP:conf/wecwis/SintorisV17,DBLP:conf/caise/0003W0LLZLW20}. For instance, researchers combined NLP with computational linguistics to generate BPMN models \cite{DBLP:conf/caise/FriedrichMP11}, used text mining to derive models directly from text \cite{DBLP:journals/jucs/GoncalvesSB11}, and applied NLP to extract structured process representations \cite{DBLP:journals/jksucis/SholiqSA22}. However, these traditional methods were often hindered by the inherent ambiguity and variability of natural language, necessitating significant human intervention and preventing full automation \cite{DBLP:conf/coling/AaCLMP18}.

The advent of LLMs has marked a paradigm shift in this domain. A significant line of research investigates the use of LLMs for generating process models directly from various inputs. Studies have demonstrated the ability of LLMs to generate process models from unstructured text \cite{DBLP:conf/bpmds/KouraniB0A24,DBLP:journals/corr/abs-2410-03255} and to translate textual descriptions into both procedural and declarative process model constraints \cite{DBLP:conf/bpm/GrohsAER23}. Beyond direct generation, other methods explore more interactive approaches, such as creating models through dialogue-based systems and chatbots \cite{DBLP:conf/bpm/KlievtsovaBKMR23}. 

\section{Evaluation Methodology}
\label{sec:methodology}
Our methodology is designed to create a controlled environment where we can systematically evaluate an LLM's tendency for knowledge-driven hallucination. The experiment consists of three main stages: (1) the generation of standard and conflicting process artifacts, (2) the procedure for generating process models using LLMs, and (3) the protocol for evaluating the generated models.

\subsection{Artifact Generation}
\label{ssec:artifacts}

We base our experiments on four diverse business processes selected from the benchmark presented in \cite{DBLP:journals/corr/abs-2412-00023}. For each of these processes, a set of ground truth artifacts already exists, which we designate as the \emph{standard} or \emph{expected} versions: a standard model ($M^+$), a corresponding natural language description ($D^+$), and a simulated event log ($L^+$). \autoref{tab:selectedprocesses} summarizes key dimensions of the selected processes. For each process, we report the number of activities and the number of nodes (transitions + places) in its ground truth Petri net. Furthermore, we indicate whether the process contains key control-flow constructs: decision points, cycles, and concurrency. 

\begin{table*}[!t]
    \caption{Characteristics of the selected processes.}

    \label{tab:selectedprocesses}
    \centering 
    \resizebox{0.8\textwidth}{!}{

    \begin{tabular}{|c|ccccc|}
    \hline
    \multirow{2}{*}{\textbf{Process}} & \multicolumn{5}{c|}{\textbf{Ground Truth Petri Net}}                               \\
                          & \#Activities & \#Nodes & Decisions & Cycles & Concurrency \\
                         \hline
    Sales Order (p1)  & 8  & 26 & $\times$ &          & $\times$ \\
    Booking System (p7)  &  13 & 49 & $\times$ & $\times$ & $\times$ \\
    Complaint Handling (p13) & 9  & 21 & $\times$ &          &          \\
    Internal Audit (p16) &  24 & 63 & $\times$ & $\times$ & $\times$ \\
    \hline
    \end{tabular}
    }
\end{table*}

From these standard artifacts, we systematically generate two sets of conflicting evidence:
\begin{itemize}
    \item \textbf{Reversed Artifacts ($M^-$, $L^-$, $D^-$):} The reversed model ($M^-$) was created by manually reversing all sequential dependencies in $M^+$. The reversed log ($L^-$) was generated by reversing the event order in each trace of $L^+$. Finally, the reversed description ($D^-$) was created by manually adjusting $D^+$ to match the new process flow of $M^-$.

    \item \textbf{Shuffled Artifacts ($M^*$, $L^*$, $D^*$):} The shuffled model ($M^*$) was created by applying a random, bijective mapping of activity labels to the standard model $M^+$, preserving its control-flow structure. The shuffled log ($L^*$) was created by applying the same mapping to the activity names in $L^+$. The corresponding description ($D^*$) was then derived manually.
\end{itemize}

This setup provides us with six distinct input scenarios for the LLM: three based on text descriptions ($D^+$, $D^-$, $D^*$) and three based on event logs ($L^+$, $L^-$, $L^*$). To compactly represent event logs as textual input for LLMs, we generate a textual abstraction for each log using the process mining library PM4Py \cite{DBLP:journals/simpa/BertiZS23}.

\begin{figure}[t!]
\centering

\begin{subfigure}{0.6\textwidth}
    \centering
\begin{subfigure}[b]{\textwidth}
        \centering
        \includegraphics[width=\linewidth]{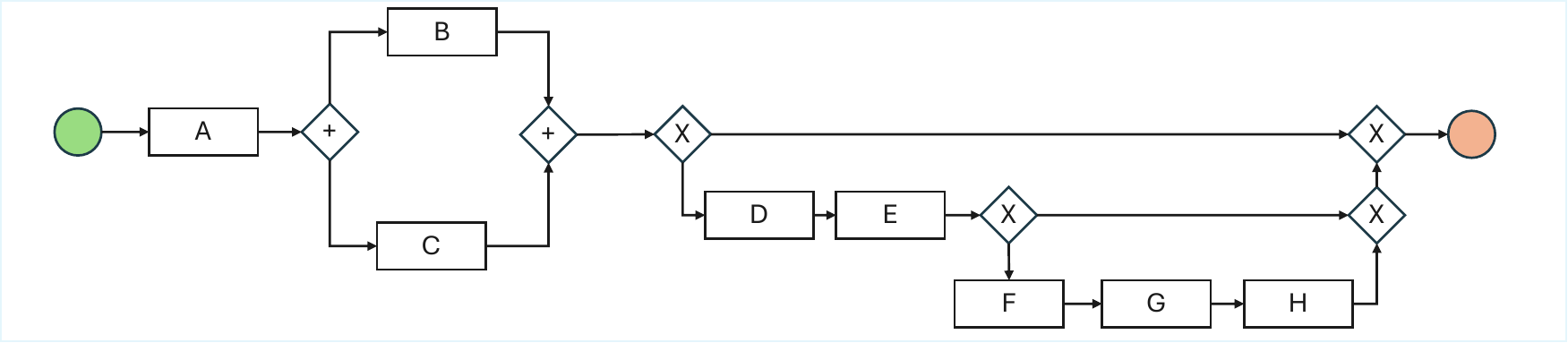}
        \caption{Original process model ($M^+$).}
    \end{subfigure}
    \vspace{0.5em} 

    \begin{subfigure}[b]{\textwidth}
        \centering
        \includegraphics[width=\textwidth]{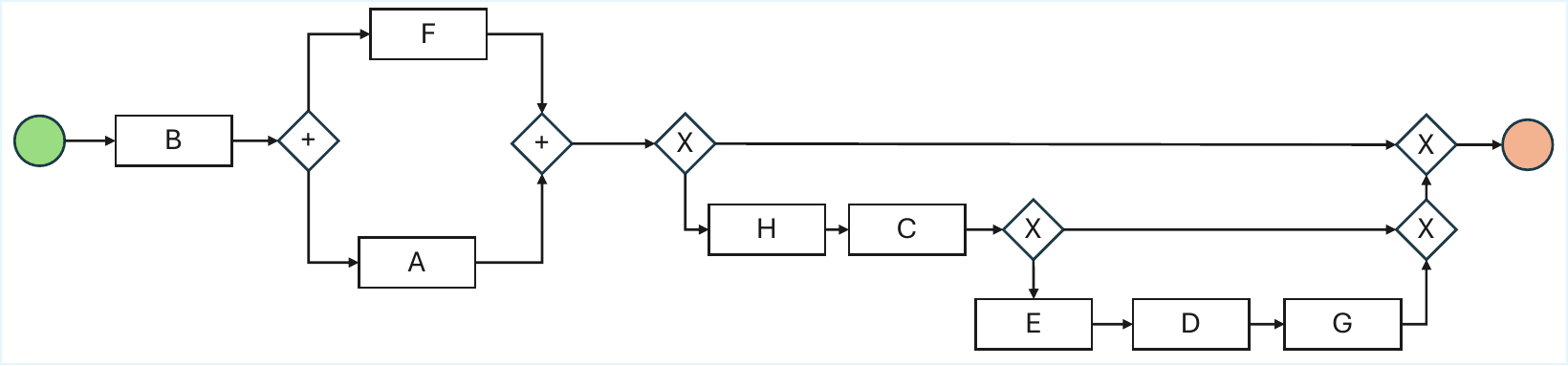}
        \caption{Shuffled process model ($M^*$).}
    \end{subfigure}

    \vspace{0.5em}

    \begin{subfigure}[b]{\textwidth}
        \centering
        \includegraphics[width=\textwidth]{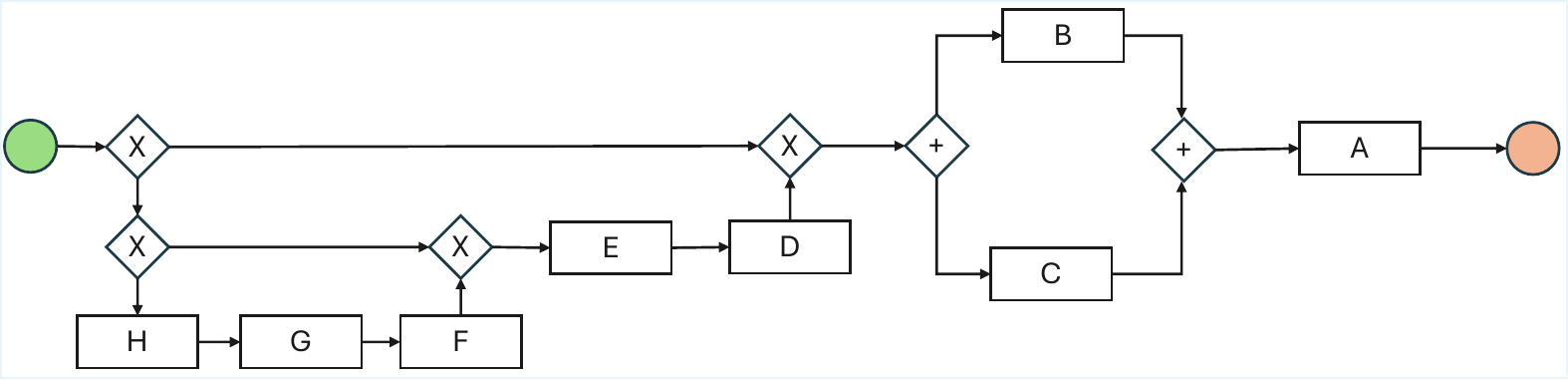}
        \caption{Reversed process model ($M^-$).}
    \end{subfigure}

\end{subfigure}
\hfill
\begin{subfigure}{0.38\textwidth}
    \centering
    
    \resizebox{\linewidth}{!}{\begin{tabular}{ll}
\toprule
\textbf{Abbreviation} & \textbf{Original Activity Label} \\
\midrule
A & Receive customer inquiry \\
B & Collect customer information \\
C & Address customer concerns or questions \\
D & Guide customer in selecting product/service \\
E & Provide quote \\
F & Place order \\
G & Record order in system \\
H & Send order confirmation to customer \\
\bottomrule
\end{tabular}}
    \caption{Mapping of abbreviations used in the figures to the original activity labels.}
    \label{fig:sales_order_models}
\end{subfigure}

\caption{Ground truth process models for the sales order process.}
\end{figure}

To illustrate the conflicting artifacts, \autoref{fig:sales_order_models} shows the standard, shuffled, and reversed models for the sales order process. The corresponding reversed description ($D^-$) and the textual abstraction of the reversed event log ($L^-$) are provided in \autoref{lst:reversedD} and \autoref{lst:reversedL}, respectively.

\begin{lstlisting}[caption={Reversed textual description ($D^-$) for the sales order process.},label={lst:reversedD}, frame=single, float, floatplacement='!t', basicstyle=\scriptsize\ttfamily]
First a confirmation of the order may be sent to the customer. If the customer
 receives a confirmation, then the order is recorded in the system and the
 order is placed. After that, the sales representative provides a quota and
 the customer is guided in selecting product or services. However, all
 previous steps can be skipped. Then, the sales staff or customer support
 addresses any concerns or questions and collects relevant information, at
 the same time. Finally, the department receives a potential customer inquiry
 about a product or service.
\end{lstlisting}

{\begin{lstlisting}[caption={The textual abstraction generated with PM4Py \cite{DBLP:journals/simpa/BertiZS23} for the reversed event log ($L^-$) for the sales order process.},label={lst:reversedL}, frame=single, float, floatplacement='!t', basicstyle=\scriptsize\ttfamily]
Send order confirmation to customer -> Record order in system -> ...
Send order confirmation to customer -> Record order in system -> ...
Provide quote -> Guide customer in selecting product/service -> ...
Provide quote -> Guide customer in selecting product/service -> ...
Collect customer information -> Address customer concerns or questions -> ...
Address customer concerns or questions -> Collect customer information -> ...
\end{lstlisting}
}

\subsection{LLM-based Model Generation Procedure}
\label{ssec:procedure}

\begin{table*}[!t]
\caption{Characteristics of the evaluated LLMs, including open-source status, reasoning capabilities, parameter estimates, announcement dates, and LiveBench 2025-07-30 \url{https://livebench.ai/} leaderboard scores.
}
\label{tab:selectedLlms}
\centering
\resizebox{0.8\textwidth}{!}{
\begin{tabular}{|l|c|c|c|c|c|}
\hline
\textbf{Model} & \textbf{Open-Source} & \textbf{Reasoning} & \textbf{Parameters} & \textbf{Announcement} & \textbf{LB Score} \\
\hline
command-r                 & $\times$ & ~        & 35B                & 2024-08-30    & \textcolor{red}{27.15} \\
gemini-2.5-flash          & ~        & $\times$ & est. 400B, 20B act.& 2025-06-17    & 64.42 \\
gemini-2.5-pro            & ~        & $\times$ & est. 1500B, 40B act.& 2025-06-17   & \textcolor{green}{69.39} \\
gpt-4.1-nano              & ~        & ~ & est. 18B, 2B act.  & 2025-04-14    & \textcolor{red}{40.51} \\
grok-3-fast               & ~        & ~        & est. 2700B, 50B act.& 2025-04-09   & 56.05 \\
grok-3-mini-fast          & ~        & $\times$        & est. 250B, 35B act.& 2025-04-09    & 62.36 \\
kimi-k2                   & $\times$ & ~        & $\approx$ 1000B, 32B act.    & 2025-07-11    & 62.70 \\
o3                        & ~        & $\times$ & est. 200B          & 2025-04-16    & \textcolor{green}{71.98} \\
o4-mini                   & ~        & $\times$ & est. 60B, 8B act.  & 2025-04-16    & 66.87 \\
qwen3-235b-a22b           & $\times$ & ~        & 235B, 22B act.     & 2025-07-25    & 64.72 \\
\hline
\end{tabular}
}
\end{table*}

The characteristics of the large language models evaluated in this study are summarized in \autoref{tab:selectedLlms}. We define two primary tasks for the LLMs:
\begin{itemize}
    \item \textbf{Text-to-Model Generation:} The LLM is prompted to generate a process model from each of the textual descriptions ($D^+$, $D^-$, $D^*$). We utilize the ProMoAI framework from \cite{DBLP:conf/ijcai/KouraniB0A24}, which generates models in the POWL language \cite{DBLP:conf/bpm/KouraniZ23} and subsequently converts them into Petri nets or BPMN diagrams for analysis.
    \item \textbf{Log-to-Model Generation:} The LLM is provided with the textual abstractions of the event logs ($L^+$, $L^-$, $L^*$) and is prompted to discover a process model that explains the behavior in each log. This experiment was also executed using the ProMoAI framework, with the event log abstraction serving as the input process description.
\end{itemize}

To investigate the LLM's sensitivity to prompting, we conducted all experiments under two distinct conditions:
\begin{itemize}
    \item \textbf{Standard Prompt:} The original, optimized prompt from ProMoAI.
    \item \textbf{Strict Adherence Prompt:} The standard prompt is adjusted with an explicit instruction for the LLM to disregard its background knowledge and fully rely on the provided input.
\end{itemize}

To ensure that a quantitative comparison is meaningful and can be fully automated, we standardize the activity labels across all experiments. For each process, the prompt provided to the LLM is extended to include the complete list of valid activity labels. This experimental design focuses the evaluation purely on the discovered control-flow structure, thereby removing any ambiguity that could arise from the LLM generating synonymous or differently phrased activity names.

\subsection{Evaluation Protocol}
\label{ssec:evaluation}

Our evaluation protocol aims to quantify the tension between evidence adherence and knowledge reversion. To measure the relationship between a generated model and the ground truth variants, we compute their \emph{semantic similarity}.
This is quantified using the behavioral-footprint similarity implemented in the PM4Py library \cite{DBLP:journals/simpa/BertiZS23}.

We acknowledge that more robust evaluation techniques, such as formal conformance checking as proposed in \cite{DBLP:journals/corr/abs-2412-00023}, exist for assessing model quality. However, the primary objective of our study is not to ascertain the absolute correctness of the generated models, but rather to perform a relative comparison. Our goal is to determine which of the three ground truth variants ($M^+$, $M^-$, or $M^*$) a generated model most closely resembles. The footprint-based similarity metric is well-suited for this purpose, as our analysis focuses on identifying the highest similarity score in each comparison rather than on the absolute values of the scores themselves.

For each LLM-generated process model, we compute its semantic similarity against all three ground truth models: $M^+$, $M^-$, and $M^*$. Ideally, a model generated from conflicting evidence (e.g., from $D^-$ or $L^-$) should exhibit high similarity to its corresponding ground truth ($M^-$). Our central hypothesis is that knowledge-driven hallucination will cause the generated model to show significant similarity to the standard model ($M^+$) instead.

\section{Results and Discussion}
\label{sec:results}
The results of our experiments are summarized\footnote{All artifacts and results
are available at \url{https://github.com/antonov1/process-hallucinations}.} in \autoref{tab:combined_text} and \autoref{tab:combined_log}. For each process model generated by an LLM, we report three semantic similarity scores, comparing it against the standard ($M^+$), reversed ($M^-$), and shuffled ($M^*$) ground truth models. To facilitate analysis, we highlight the highest similarity score for each generated model. The cell is colored green if the highest score corresponds to the correct ground truth artifact (e.g., a model from $D^-$ is most similar to $M^-$), indicating successful adherence to the provided evidence. Conversely, the cell is colored red if the generated model is most similar to the standard process model ($M^+$) despite being generated from conflicting evidence ($D^-$, $D^*$, $L^-$, or $L^*$), signaling a clear instance of knowledge-driven hallucination. We omit this highlighting in cases where all similarity scores for a given model are below $0.1$, as a comparison at such a low level of quality is no longer insightful.

\begin{table}[!t]
\caption{Semantic similarity scores for models generated from textual descriptions using standard and strict adherence prompts.}
\label{tab:combined_text}
    \centering
    \tiny
    \resizebox{1.0\textwidth}{!}{
    \begin{tabular}{|l|l|rrr|rrr|rrr|rrr|r|r||rrr|rrr|rrr|rrr|r|r|}
        \hline
        \multirow{3}{*}{LLM} & \multirow{3}{*}{} & \multicolumn{14}{c||}{Standard Prompt} & \multicolumn{14}{c|}{Strict Adherence Prompt} \\
        \cline{3-30}
         &  & \multicolumn{3}{c|}{Sales Order} & \multicolumn{3}{c|}{Booking} & \multicolumn{3}{c|}{Complaint} & \multicolumn{3}{c|}{Audit} & \multicolumn{2}{c|}{$\overline{diag}$} & \multicolumn{3}{c|}{Sales Order} & \multicolumn{3}{c|}{Booking} & \multicolumn{3}{c|}{Complaint} & \multicolumn{3}{c|}{Audit} & \multicolumn{2}{c|}{$\overline{diag}$} \\
        \cline{3-14} \cline{15-16} \cline{17-28} \cline{29-30}
         &  & $M^+$ & $M^-$ & $M^*$ & $M^+$ & $M^-$ & $M^*$ & $M^+$ & $M^-$ & $M^*$ & $M^+$ & $M^-$ & $M^*$ &  & All & $M^+$ & $M^-$ & $M^*$ & $M^+$ & $M^-$ & $M^*$ & $M^+$ & $M^-$ & $M^*$ & $M^+$ & $M^-$ & $M^*$ &  & All \\
        \hline

        \multirow{3}{*}{command-r}& $D^{+}$  & \cellcolor[HTML]{B7E4B7} 0.20 & 0.05 & 0.00 & \cellcolor[HTML]{B7E4B7} 0.37 & 0.06 & 0.06 & 0.04 & 0.00 & 0.01 & \cellcolor[HTML]{B7E4B7} 0.17 & 0.09 & 0.05 & 0.20 & \multirow{3}{*}{0.18} & 0.10 & 0.00 & 0.00 & 0.10 & 0.06 & 0.05 & \cellcolor[HTML]{B7E4B7} 0.19 & 0.00 & 0.00 & \cellcolor[HTML]{B7E4B7} 0.12 & 0.07 & 0.02 & 0.13 & \multirow{3}{*}{0.17} \\
        & $D^{-}$  & 0.03 & 0.00 & 0.03 & \cellcolor[HTML]{F4CCCC} 0.12 & 0.02 & 0.02 & 0.00 & \cellcolor[HTML]{B7E4B7} 0.90 & 0.06 & 0.12 & \cellcolor[HTML]{B7E4B7} 0.23 & 0.04 & 0.29 &  & 0.10 & \cellcolor[HTML]{B7E4B7} 0.16 & 0.05 & \cellcolor[HTML]{F4CCCC} 0.12 & 0.02 & 0.02 & 0.00 & \cellcolor[HTML]{B7E4B7} 0.58 & 0.00 & 0.12 & \cellcolor[HTML]{B7E4B7} 0.23 & 0.04 & 0.25 &  \\
        & $D^{*}$  & \cellcolor[HTML]{F4CCCC} 0.24 & 0.17 & 0.05 & 0.04 & 0.00 & 0.06 & 0.01 & 0.00 & 0.04 & 0.04 & 0.03 & 0.05 & 0.05 &  & \cellcolor[HTML]{F4CCCC} 0.19 & 0.06 & 0.12 & 0.04 & 0.00 & 0.06 & 0.05 & 0.05 & \cellcolor[HTML]{B7E4B7} 0.33 & 0.02 & 0.02 & 0.02 & 0.13 &  \\
        \hline

        \multirow{3}{*}{gemini-2.5-flash}& $D^{+}$  & \cellcolor[HTML]{B7E4B7} 0.40 & 0.08 & 0.04 & \cellcolor[HTML]{B7E4B7} 0.48 & 0.11 & 0.03 & \cellcolor[HTML]{B7E4B7} 1.00 & 0.00 & 0.06 & \cellcolor[HTML]{B7E4B7} 0.53 & 0.26 & 0.02 & 0.60 & \multirow{3}{*}{0.45} & \cellcolor[HTML]{B7E4B7} 0.40 & 0.08 & 0.04 & \cellcolor[HTML]{B7E4B7} 0.48 & 0.11 & 0.03 & \cellcolor[HTML]{B7E4B7} 0.62 & 0.00 & 0.11 & \cellcolor[HTML]{B7E4B7} 0.65 & 0.22 & 0.03 & 0.54 & \multirow{3}{*}{0.46} \\
        & $D^{-}$  & 0.09 & \cellcolor[HTML]{B7E4B7} 0.60 & 0.04 & \cellcolor[HTML]{F4CCCC} 0.59 & 0.23 & 0.05 & 0.00 & \cellcolor[HTML]{B7E4B7} 1.00 & 0.06 & 0.07 & \cellcolor[HTML]{B7E4B7} 0.17 & 0.01 & 0.50 &  & 0.09 & \cellcolor[HTML]{B7E4B7} 0.60 & 0.04 & 0.10 & 0.08 & 0.06 & 0.00 & \cellcolor[HTML]{B7E4B7} 1.00 & 0.06 & 0.14 & \cellcolor[HTML]{B7E4B7} 0.29 & 0.01 & 0.49 &  \\
        & $D^{*}$  & 0.05 & 0.05 & \cellcolor[HTML]{B7E4B7} 0.33 & 0.05 & 0.06 & \cellcolor[HTML]{B7E4B7} 0.29 & \cellcolor[HTML]{F4CCCC} 0.58 & 0.06 & 0.12 & 0.07 & 0.04 & \cellcolor[HTML]{B7E4B7} 0.21 & 0.24 &  & 0.08 & 0.04 & \cellcolor[HTML]{B7E4B7} 0.40 & 0.03 & 0.06 & \cellcolor[HTML]{B7E4B7} 0.35 & 0.14 & 0.00 & \cellcolor[HTML]{B7E4B7} 0.39 & 0.05 & 0.03 & \cellcolor[HTML]{B7E4B7} 0.23 & 0.34 &  \\
        \hline

        \multirow{3}{*}{gemini-2.5-pro}& $D^{+}$  & \cellcolor[HTML]{B7E4B7} 1.00 & 0.11 & 0.00 & \cellcolor[HTML]{B7E4B7} 0.52 & 0.13 & 0.04 & \cellcolor[HTML]{B7E4B7} 1.00 & 0.00 & 0.06 & \cellcolor[HTML]{B7E4B7} 0.68 & 0.22 & 0.02 & \textcolor{darkgreen}{\textbf{0.80}} & \multirow{3}{*}{0.41} & \cellcolor[HTML]{B7E4B7} 1.00 & 0.11 & 0.00 & \cellcolor[HTML]{B7E4B7} 0.41 & 0.12 & 0.06 & \cellcolor[HTML]{B7E4B7} 0.90 & 0.00 & 0.06 & \cellcolor[HTML]{B7E4B7} 0.77 & 0.23 & 0.02 & \textcolor{darkgreen}{\textbf{0.77}} & \multirow{3}{*}{0.47} \\
        & $D^{-}$  & \cellcolor[HTML]{F4CCCC} 0.44 & 0.08 & 0.08 & \cellcolor[HTML]{F4CCCC} 0.38 & 0.13 & 0.05 & 0.00 & \cellcolor[HTML]{B7E4B7} 1.00 & 0.06 & \cellcolor[HTML]{F4CCCC} 0.19 & 0.12 & 0.01 & 0.33 &  & 0.08 & \cellcolor[HTML]{B7E4B7} 0.40 & 0.04 & \cellcolor[HTML]{F4CCCC} 0.15 & 0.10 & 0.08 & 0.00 & \cellcolor[HTML]{B7E4B7} 1.00 & 0.06 & 0.14 & \cellcolor[HTML]{B7E4B7} 0.17 & 0.01 & 0.42 &  \\
        & $D^{*}$  & \cellcolor[HTML]{F4CCCC} 0.35 & 0.08 & 0.08 & \cellcolor[HTML]{F4CCCC} 0.27 & 0.09 & 0.06 & \cellcolor[HTML]{F4CCCC} 0.80 & 0.00 & 0.12 & 0.08 & 0.06 & \cellcolor[HTML]{B7E4B7} 0.17 & 0.11 &  & 0.07 & 0.04 & \cellcolor[HTML]{B7E4B7} 0.32 & 0.09 & 0.09 & \cellcolor[HTML]{B7E4B7} 0.16 & \cellcolor[HTML]{F4CCCC} 0.58 & 0.06 & 0.12 & 0.04 & 0.04 & \cellcolor[HTML]{B7E4B7} 0.30 & 0.22 &  \\
        \hline

        \multirow{3}{*}{gpt-4.1-nano}& $D^{+}$  & \cellcolor[HTML]{B7E4B7} 0.46 & 0.00 & 0.00 & 0.01 & 0.00 & 0.01 & \cellcolor[HTML]{B7E4B7} 0.69 & 0.00 & 0.05 & 0.01 & 0.00 & 0.01 & 0.29 & \multirow{3}{*}{0.23} & \cellcolor[HTML]{B7E4B7} 0.91 & 0.11 & 0.00 & \cellcolor[HTML]{B7E4B7} 0.23 & 0.07 & 0.04 & 0.08 & 0.00 & 0.00 & 0.06 & 0.04 & 0.02 & 0.32 & \multirow{3}{*}{0.22} \\
        & $D^{-}$  & 0.14 & 0.00 & 0.14 & \cellcolor[HTML]{F4CCCC} 0.34 & 0.07 & 0.00 & 0.00 & \cellcolor[HTML]{B7E4B7} 0.73 & 0.06 & 0.04 & 0.04 & 0.01 & 0.21 &  & \cellcolor[HTML]{F4CCCC} 0.27 & 0.00 & 0.15 & \cellcolor[HTML]{F4CCCC} 0.38 & 0.10 & 0.02 & 0.00 & \cellcolor[HTML]{B7E4B7} 0.50 & 0.06 & 0.19 & \cellcolor[HTML]{B7E4B7} 0.37 & 0.02 & 0.24 &  \\
        & $D^{*}$  & \cellcolor[HTML]{F4CCCC} 0.33 & 0.07 & 0.07 & 0.05 & 0.05 & \cellcolor[HTML]{B7E4B7} 0.25 & 0.19 & 0.00 & \cellcolor[HTML]{B7E4B7} 0.36 & 0.08 & 0.02 & 0.06 & 0.18 &  & \cellcolor[HTML]{F4CCCC} 0.33 & 0.00 & 0.07 & 0.02 & 0.06 & \cellcolor[HTML]{B7E4B7} 0.12 & \cellcolor[HTML]{F4CCCC} 0.36 & 0.00 & 0.25 & 0.00 & 0.00 & 0.00 & 0.11 &  \\
        \hline

        \multirow{3}{*}{grok-3-fast}& $D^{+}$  & \cellcolor[HTML]{B7E4B7} 1.00 & 0.11 & 0.00 & \cellcolor[HTML]{B7E4B7} 0.21 & 0.07 & 0.04 & \cellcolor[HTML]{B7E4B7} 1.00 & 0.00 & 0.06 & \cellcolor[HTML]{B7E4B7} 0.42 & 0.21 & 0.00 & 0.66 & \multirow{3}{*}{0.50} & \cellcolor[HTML]{B7E4B7} 0.55 & 0.00 & 0.00 & \cellcolor[HTML]{B7E4B7} 0.20 & 0.06 & 0.06 & \cellcolor[HTML]{B7E4B7} 1.00 & 0.00 & 0.06 & \cellcolor[HTML]{B7E4B7} 0.45 & 0.22 & 0.01 & 0.55 & \multirow{3}{*}{0.55} \\
        & $D^{-}$  & 0.11 & \cellcolor[HTML]{B7E4B7} 0.67 & 0.11 & \cellcolor[HTML]{F4CCCC} 0.66 & 0.15 & 0.02 & 0.00 & \cellcolor[HTML]{B7E4B7} 1.00 & 0.06 & 0.17 & \cellcolor[HTML]{B7E4B7} 0.39 & 0.03 & 0.55 &  & 0.09 & \cellcolor[HTML]{B7E4B7} 0.60 & 0.04 & \cellcolor[HTML]{F4CCCC} 0.38 & 0.08 & 0.00 & 0.00 & \cellcolor[HTML]{B7E4B7} 1.00 & 0.06 & 0.15 & \cellcolor[HTML]{B7E4B7} 0.29 & 0.03 & 0.49 &  \\
        & $D^{*}$  & 0.00 & 0.12 & \cellcolor[HTML]{B7E4B7} 0.50 & 0.05 & 0.09 & \cellcolor[HTML]{B7E4B7} 0.14 & \cellcolor[HTML]{F4CCCC} 0.54 & 0.05 & 0.11 & 0.05 & 0.03 & \cellcolor[HTML]{B7E4B7} 0.45 & 0.30 &  & 0.00 & 0.11 & \cellcolor[HTML]{B7E4B7} 1.00 & 0.04 & 0.09 & \cellcolor[HTML]{B7E4B7} 0.23 & 0.06 & 0.06 & \cellcolor[HTML]{B7E4B7} 0.80 & 0.06 & 0.03 & \cellcolor[HTML]{B7E4B7} 0.38 & \textcolor{darkgreen}{\textbf{0.60}} &  \\
        \hline

        \multirow{3}{*}{grok-3-mini-fast}& $D^{+}$  & \cellcolor[HTML]{B7E4B7} 0.55 & 0.00 & 0.00 & \cellcolor[HTML]{B7E4B7} 0.29 & 0.08 & 0.04 & \cellcolor[HTML]{B7E4B7} 1.00 & 0.00 & 0.06 & 0.08 & 0.05 & 0.04 & 0.48 & \multirow{3}{*}{0.41} & \cellcolor[HTML]{B7E4B7} 0.55 & 0.00 & 0.00 & \cellcolor[HTML]{B7E4B7} 0.38 & 0.08 & 0.00 & \cellcolor[HTML]{B7E4B7} 1.00 & 0.00 & 0.06 & \cellcolor[HTML]{B7E4B7} 0.49 & 0.17 & 0.02 & 0.60 & \multirow{3}{*}{0.58} \\
        & $D^{-}$  & 0.11 & \cellcolor[HTML]{B7E4B7} 1.00 & 0.11 & \cellcolor[HTML]{F4CCCC} 0.24 & 0.14 & 0.08 & 0.00 & \cellcolor[HTML]{B7E4B7} 1.00 & 0.06 & 0.16 & \cellcolor[HTML]{B7E4B7} 0.33 & 0.01 & 0.62 &  & 0.11 & \cellcolor[HTML]{B7E4B7} 1.00 & 0.11 & \cellcolor[HTML]{F4CCCC} 0.17 & 0.00 & 0.04 & 0.00 & \cellcolor[HTML]{B7E4B7} 1.00 & 0.06 & 0.16 & \cellcolor[HTML]{B7E4B7} 0.33 & 0.01 & 0.58 &  \\
        & $D^{*}$  & \cellcolor[HTML]{F4CCCC} 0.55 & 0.00 & 0.00 & 0.08 & 0.07 & \cellcolor[HTML]{B7E4B7} 0.15 & \cellcolor[HTML]{F4CCCC} 0.55 & 0.00 & 0.13 & 0.02 & 0.03 & \cellcolor[HTML]{B7E4B7} 0.30 & 0.14 &  & 0.05 & 0.10 & \cellcolor[HTML]{B7E4B7} 0.83 & 0.05 & 0.06 & \cellcolor[HTML]{B7E4B7} 0.24 & 0.00 & 0.06 & \cellcolor[HTML]{B7E4B7} 0.70 & 0.03 & 0.03 & \cellcolor[HTML]{B7E4B7} 0.42 & 0.55 &  \\
        \hline

        \multirow{3}{*}{kimi-k2}& $D^{+}$  & \cellcolor[HTML]{B7E4B7} 0.44 & 0.08 & 0.00 & \cellcolor[HTML]{B7E4B7} 0.54 & 0.11 & 0.00 & \cellcolor[HTML]{B7E4B7} 0.47 & 0.00 & 0.09 & \cellcolor[HTML]{B7E4B7} 0.38 & 0.15 & 0.02 & 0.46 & \multirow{3}{*}{0.40} & \cellcolor[HTML]{B7E4B7} 1.00 & 0.11 & 0.00 & \cellcolor[HTML]{B7E4B7} 0.54 & 0.11 & 0.00 & \cellcolor[HTML]{B7E4B7} 1.00 & 0.00 & 0.06 & \cellcolor[HTML]{B7E4B7} 0.47 & 0.23 & 0.02 & 0.75 & \multirow{3}{*}{0.44} \\
        & $D^{-}$  & 0.24 & \cellcolor[HTML]{B7E4B7} 0.62 & 0.11 & \cellcolor[HTML]{F4CCCC} 0.38 & 0.08 & 0.00 & 0.00 & \cellcolor[HTML]{B7E4B7} 0.90 & 0.06 & 0.12 & \cellcolor[HTML]{B7E4B7} 0.54 & 0.03 & 0.53 &  & 0.24 & \cellcolor[HTML]{B7E4B7} 0.62 & 0.11 & \cellcolor[HTML]{F4CCCC} 0.22 & 0.09 & 0.04 & 0.00 & \cellcolor[HTML]{B7E4B7} 0.14 & 0.05 & 0.11 & \cellcolor[HTML]{B7E4B7} 0.44 & 0.03 & 0.32 &  \\
        & $D^{*}$  & 0.04 & 0.00 & \cellcolor[HTML]{B7E4B7} 0.19 & 0.05 & 0.05 & \cellcolor[HTML]{B7E4B7} 0.21 & \cellcolor[HTML]{F4CCCC} 0.27 & 0.03 & 0.06 & 0.04 & 0.03 & \cellcolor[HTML]{B7E4B7} 0.31 & 0.19 &  & 0.05 & 0.05 & \cellcolor[HTML]{B7E4B7} 0.33 & 0.05 & 0.07 & \cellcolor[HTML]{B7E4B7} 0.15 & \cellcolor[HTML]{F4CCCC} 0.33 & 0.00 & 0.23 & 0.04 & 0.01 & \cellcolor[HTML]{B7E4B7} 0.26 & 0.24 &  \\
        \hline

        \multirow{3}{*}{o3}& $D^{+}$  & \cellcolor[HTML]{B7E4B7} 0.50 & 0.00 & 0.06 & \cellcolor[HTML]{B7E4B7} 0.34 & 0.09 & 0.03 & \cellcolor[HTML]{B7E4B7} 0.90 & 0.00 & 0.06 & \cellcolor[HTML]{B7E4B7} 0.66 & 0.22 & 0.01 & 0.60 & \multirow{3}{*}{0.42} & \cellcolor[HTML]{B7E4B7} 1.00 & 0.11 & 0.00 & \cellcolor[HTML]{B7E4B7} 0.26 & 0.08 & 0.03 & \cellcolor[HTML]{B7E4B7} 0.90 & 0.00 & 0.06 & \cellcolor[HTML]{B7E4B7} 0.75 & 0.24 & 0.01 & 0.73 & \multirow{3}{*}{0.61} \\
        & $D^{-}$  & 0.25 & 0.25 & 0.25 & \cellcolor[HTML]{F4CCCC} 0.45 & 0.19 & 0.03 & 0.00 & \cellcolor[HTML]{B7E4B7} 1.00 & 0.06 & \cellcolor[HTML]{F4CCCC} 0.27 & 0.24 & 0.02 & 0.42 &  & 0.11 & \cellcolor[HTML]{B7E4B7} 1.00 & 0.11 & 0.13 & \cellcolor[HTML]{B7E4B7} 0.14 & 0.06 & 0.00 & \cellcolor[HTML]{B7E4B7} 1.00 & 0.06 & 0.09 & \cellcolor[HTML]{B7E4B7} 0.18 & 0.05 & 0.58 &  \\
        & $D^{*}$  & 0.06 & 0.06 & \cellcolor[HTML]{B7E4B7} 0.50 & 0.07 & 0.05 & \cellcolor[HTML]{B7E4B7} 0.14 & \cellcolor[HTML]{F4CCCC} 0.58 & 0.06 & 0.12 & 0.04 & 0.02 & \cellcolor[HTML]{B7E4B7} 0.24 & 0.25 &  & 0.05 & 0.10 & \cellcolor[HTML]{B7E4B7} 0.83 & 0.05 & 0.05 & \cellcolor[HTML]{B7E4B7} 0.19 & 0.00 & 0.06 & \cellcolor[HTML]{B7E4B7} 0.64 & 0.03 & 0.02 & \cellcolor[HTML]{B7E4B7} 0.44 & 0.53 &  \\
        \hline

        \multirow{3}{*}{o4-mini}& $D^{+}$  & \cellcolor[HTML]{B7E4B7} 0.55 & 0.00 & 0.00 & \cellcolor[HTML]{B7E4B7} 0.31 & 0.09 & 0.03 & \cellcolor[HTML]{B7E4B7} 0.90 & 0.00 & 0.06 & \cellcolor[HTML]{B7E4B7} 0.35 & 0.19 & 0.03 & 0.53 & \multirow{3}{*}{\textcolor{darkgreen}{\textbf{0.55}}} & \cellcolor[HTML]{B7E4B7} 1.00 & 0.11 & 0.00 & \cellcolor[HTML]{B7E4B7} 0.41 & 0.12 & 0.04 & \cellcolor[HTML]{B7E4B7} 0.90 & 0.00 & 0.06 & \cellcolor[HTML]{B7E4B7} 0.54 & 0.26 & 0.02 & 0.71 & \multirow{3}{*}{\textcolor{darkgreen}{\textbf{0.63}}} \\
        & $D^{-}$  & 0.11 & \cellcolor[HTML]{B7E4B7} 1.00 & 0.11 & \cellcolor[HTML]{F4CCCC} 0.66 & 0.15 & 0.03 & 0.00 & \cellcolor[HTML]{B7E4B7} 1.00 & 0.06 & 0.30 & \cellcolor[HTML]{B7E4B7} 0.40 & 0.02 & \textcolor{darkgreen}{\textbf{0.64}} &  & 0.11 & \cellcolor[HTML]{B7E4B7} 1.00 & 0.11 & 0.17 & \cellcolor[HTML]{B7E4B7} 0.19 & 0.05 & 0.00 & \cellcolor[HTML]{B7E4B7} 1.00 & 0.06 & 0.15 & \cellcolor[HTML]{B7E4B7} 0.40 & 0.02 & 0.65 &  \\
        & $D^{*}$  & 0.05 & 0.10 & \cellcolor[HTML]{B7E4B7} 0.64 & 0.05 & 0.06 & \cellcolor[HTML]{B7E4B7} 0.26 & 0.06 & 0.06 & \cellcolor[HTML]{B7E4B7} 0.80 & 0.03 & 0.03 & \cellcolor[HTML]{B7E4B7} 0.28 & \textcolor{darkgreen}{\textbf{0.50}} &  & 0.05 & 0.10 & \cellcolor[HTML]{B7E4B7} 0.64 & 0.05 & 0.06 & \cellcolor[HTML]{B7E4B7} 0.26 & 0.06 & 0.06 & \cellcolor[HTML]{B7E4B7} 0.80 & 0.03 & 0.04 & \cellcolor[HTML]{B7E4B7} 0.44 & 0.54 &  \\
        \hline

        \multirow{3}{*}{qwen3-235b-a22b}& $D^{+}$  & \cellcolor[HTML]{B7E4B7} 1.00 & 0.11 & 0.00 & \cellcolor[HTML]{B7E4B7} 0.30 & 0.09 & 0.03 & \cellcolor[HTML]{B7E4B7} 1.00 & 0.00 & 0.06 & \cellcolor[HTML]{B7E4B7} 0.64 & 0.24 & 0.01 & 0.73 & \multirow{3}{*}{0.45} & \cellcolor[HTML]{B7E4B7} 0.50 & 0.00 & 0.06 & \cellcolor[HTML]{B7E4B7} 0.52 & 0.13 & 0.04 & \cellcolor[HTML]{B7E4B7} 0.57 & 0.00 & 0.10 & \cellcolor[HTML]{B7E4B7} 0.25 & 0.16 & 0.03 & 0.46 & \multirow{3}{*}{0.52} \\
        & $D^{-}$  & 0.24 & \cellcolor[HTML]{B7E4B7} 0.62 & 0.11 & \cellcolor[HTML]{F4CCCC} 0.66 & 0.15 & 0.02 & 0.00 & \cellcolor[HTML]{B7E4B7} 1.00 & 0.06 & \cellcolor[HTML]{F4CCCC} 0.26 & 0.23 & 0.03 & 0.50 &  & 0.11 & \cellcolor[HTML]{B7E4B7} 1.00 & 0.11 & \cellcolor[HTML]{F4CCCC} 0.34 & 0.11 & 0.04 & 0.00 & \cellcolor[HTML]{B7E4B7} 1.00 & 0.06 & 0.16 & \cellcolor[HTML]{B7E4B7} 0.51 & 0.04 & \textcolor{darkgreen}{\textbf{0.66}} &  \\
        & $D^{*}$  & 0.10 & 0.00 & 0.10 & \cellcolor[HTML]{F4CCCC} 0.13 & 0.04 & 0.03 & \cellcolor[HTML]{F4CCCC} 0.73 & 0.06 & 0.06 & 0.04 & 0.01 & \cellcolor[HTML]{B7E4B7} 0.28 & 0.11 &  & 0.00 & 0.06 & \cellcolor[HTML]{B7E4B7} 0.46 & 0.03 & 0.06 & \cellcolor[HTML]{B7E4B7} 0.35 & 0.20 & 0.00 & \cellcolor[HTML]{B7E4B7} 0.64 & 0.05 & 0.03 & \cellcolor[HTML]{B7E4B7} 0.29 & 0.43 &  \\
        \hline

    \end{tabular}
    }

\end{table}

\begin{table}[!t]
\caption{Semantic similarity scores for models generated from event logs using standard and strict adherence prompts.}
\label{tab:combined_log}
    \centering
    \tiny
    \resizebox{1.0\textwidth}{!}{
    \begin{tabular}{|l|l|rrr|rrr|rrr|rrr|r|r||rrr|rrr|rrr|rrr|r|r|}
        \hline
        \multirow{3}{*}{LLM} & \multirow{3}{*}{} & \multicolumn{14}{c||}{Standard Prompt} & \multicolumn{14}{c|}{Strict Adherence Prompt} \\
        \cline{3-30}
         &  & \multicolumn{3}{c|}{Sales Order} & \multicolumn{3}{c|}{Booking} & \multicolumn{3}{c|}{Complaint} & \multicolumn{3}{c|}{Audit} & \multicolumn{2}{c|}{$\overline{diag}$} & \multicolumn{3}{c|}{Sales Order} & \multicolumn{3}{c|}{Booking} & \multicolumn{3}{c|}{Complaint} & \multicolumn{3}{c|}{Audit} & \multicolumn{2}{c|}{$\overline{diag}$} \\
        \cline{3-14} \cline{15-16} \cline{17-28} \cline{29-30}
         &  & $M^+$ & $M^-$ & $M^*$ & $M^+$ & $M^-$ & $M^*$ & $M^+$ & $M^-$ & $M^*$ & $M^+$ & $M^-$ & $M^*$ &  & All & $M^+$ & $M^-$ & $M^*$ & $M^+$ & $M^-$ & $M^*$ & $M^+$ & $M^-$ & $M^*$ & $M^+$ & $M^-$ & $M^*$ &  & All \\
        \hline

        \multirow{3}{*}{command-r}& $L^{+}$  & \cellcolor[HTML]{B7E4B7} 0.60 & 0.09 & 0.04 & \cellcolor[HTML]{B7E4B7} 0.12 & 0.03 & 0.03 & \cellcolor[HTML]{B7E4B7} 0.80 & 0.00 & 0.06 & 0.03 & 0.03 & 0.03 & 0.39 & \multirow{3}{*}{0.26} & \cellcolor[HTML]{B7E4B7} 0.60 & 0.09 & 0.04 & \cellcolor[HTML]{B7E4B7} 0.12 & 0.03 & 0.03 & \cellcolor[HTML]{B7E4B7} 0.80 & 0.00 & 0.06 & 0.03 & 0.03 & 0.03 & 0.39 & \multirow{3}{*}{0.30} \\
        & $L^{-}$  & 0.08 & \cellcolor[HTML]{B7E4B7} 0.42 & 0.08 & 0.00 & \cellcolor[HTML]{B7E4B7} 0.19 & 0.02 & 0.00 & 0.00 & 0.00 & 0.04 & \cellcolor[HTML]{B7E4B7} 0.27 & 0.01 & 0.22 &  & 0.06 & \cellcolor[HTML]{B7E4B7} 0.36 & 0.12 & 0.00 & \cellcolor[HTML]{B7E4B7} 0.19 & 0.02 & 0.00 & \cellcolor[HTML]{B7E4B7} 0.62 & 0.05 & 0.04 & \cellcolor[HTML]{B7E4B7} 0.27 & 0.01 & 0.36 &  \\
        & $L^{*}$  & 0.00 & 0.06 & \cellcolor[HTML]{B7E4B7} 0.42 & 0.04 & 0.04 & \cellcolor[HTML]{B7E4B7} 0.11 & 0.11 & 0.00 & \cellcolor[HTML]{B7E4B7} 0.18 & 0.00 & 0.00 & 0.00 & 0.18 &  & 0.05 & 0.05 & \cellcolor[HTML]{B7E4B7} 0.33 & 0.04 & 0.03 & \cellcolor[HTML]{B7E4B7} 0.11 & 0.00 & 0.02 & \cellcolor[HTML]{B7E4B7} 0.15 & 0.00 & 0.00 & 0.00 & 0.15 &  \\
        \hline

        \multirow{3}{*}{gemini-2.5-flash}& $L^{+}$  & \cellcolor[HTML]{B7E4B7} 1.00 & 0.11 & 0.00 & \cellcolor[HTML]{B7E4B7} 0.75 & 0.19 & 0.03 & \cellcolor[HTML]{B7E4B7} 1.00 & 0.00 & 0.06 & \cellcolor[HTML]{B7E4B7} 0.68 & 0.18 & 0.05 & \textcolor{darkgreen}{\textbf{0.86}} & \multirow{3}{*}{0.57} & \cellcolor[HTML]{B7E4B7} 1.00 & 0.11 & 0.00 & \cellcolor[HTML]{B7E4B7} 0.62 & 0.15 & 0.02 & \cellcolor[HTML]{B7E4B7} 1.00 & 0.00 & 0.06 & \cellcolor[HTML]{B7E4B7} 0.67 & 0.18 & 0.04 & 0.82 & \multirow{3}{*}{0.70} \\
        & $L^{-}$  & \cellcolor[HTML]{F4CCCC} 1.00 & 0.11 & 0.00 & \cellcolor[HTML]{F4CCCC} 0.91 & 0.22 & 0.05 & 0.20 & \cellcolor[HTML]{B7E4B7} 0.29 & 0.06 & \cellcolor[HTML]{F4CCCC} 0.47 & 0.18 & 0.01 & 0.20 &  & 0.11 & \cellcolor[HTML]{B7E4B7} 1.00 & 0.11 & \cellcolor[HTML]{F4CCCC} 0.91 & 0.22 & 0.05 & 0.00 & \cellcolor[HTML]{B7E4B7} 1.00 & 0.06 & \cellcolor[HTML]{F4CCCC} 0.62 & 0.20 & 0.02 & 0.61 &  \\
        & $L^{*}$  & 0.00 & 0.10 & \cellcolor[HTML]{B7E4B7} 0.69 & 0.03 & 0.05 & \cellcolor[HTML]{B7E4B7} 0.62 & 0.06 & 0.06 & \cellcolor[HTML]{B7E4B7} 1.00 & 0.06 & 0.03 & \cellcolor[HTML]{B7E4B7} 0.32 & 0.66 &  & 0.00 & 0.11 & \cellcolor[HTML]{B7E4B7} 1.00 & 0.09 & 0.11 & \cellcolor[HTML]{B7E4B7} 0.35 & 0.06 & 0.06 & \cellcolor[HTML]{B7E4B7} 1.00 & 0.02 & 0.01 & \cellcolor[HTML]{B7E4B7} 0.34 & 0.67 &  \\
        \hline

        \multirow{3}{*}{gemini-2.5-pro}& $L^{+}$  & \cellcolor[HTML]{B7E4B7} 1.00 & 0.11 & 0.00 & \cellcolor[HTML]{B7E4B7} 0.75 & 0.19 & 0.03 & \cellcolor[HTML]{B7E4B7} 1.00 & 0.00 & 0.06 & \cellcolor[HTML]{B7E4B7} 0.68 & 0.18 & 0.05 & \textcolor{darkgreen}{\textbf{0.86}} & \multirow{3}{*}{0.59} & \cellcolor[HTML]{B7E4B7} 1.00 & 0.11 & 0.00 & \cellcolor[HTML]{B7E4B7} 0.75 & 0.19 & 0.03 & \cellcolor[HTML]{B7E4B7} 1.00 & 0.00 & 0.06 & \cellcolor[HTML]{B7E4B7} 0.68 & 0.18 & 0.05 & 0.86 & \multirow{3}{*}{0.59} \\
        & $L^{-}$  & \cellcolor[HTML]{F4CCCC} 1.00 & 0.11 & 0.00 & \cellcolor[HTML]{F4CCCC} 0.91 & 0.22 & 0.05 & \cellcolor[HTML]{F4CCCC} 1.00 & 0.00 & 0.06 & \cellcolor[HTML]{F4CCCC} 0.68 & 0.15 & 0.02 & 0.12 &  & \cellcolor[HTML]{F4CCCC} 1.00 & 0.11 & 0.00 & \cellcolor[HTML]{F4CCCC} 0.91 & 0.22 & 0.05 & 0.00 & \cellcolor[HTML]{B7E4B7} 1.00 & 0.06 & \cellcolor[HTML]{F4CCCC} 0.68 & 0.15 & 0.03 & 0.37 &  \\
        & $L^{*}$  & 0.00 & 0.11 & \cellcolor[HTML]{B7E4B7} 1.00 & 0.03 & 0.03 & \cellcolor[HTML]{B7E4B7} 0.75 & 0.06 & 0.06 & \cellcolor[HTML]{B7E4B7} 1.00 & 0.03 & 0.03 & \cellcolor[HTML]{B7E4B7} 0.47 & \textcolor{darkgreen}{\textbf{0.80}} &  & 0.04 & 0.00 & \cellcolor[HTML]{B7E4B7} 0.19 & 0.02 & 0.03 & \cellcolor[HTML]{B7E4B7} 0.60 & 0.06 & 0.06 & \cellcolor[HTML]{B7E4B7} 1.00 & 0.06 & 0.07 & \cellcolor[HTML]{B7E4B7} 0.41 & 0.55 &  \\
        \hline

        \multirow{3}{*}{gpt-4.1-nano}& $L^{+}$  & \cellcolor[HTML]{B7E4B7} 0.83 & 0.10 & 0.00 & 0.03 & 0.00 & 0.00 & \cellcolor[HTML]{B7E4B7} 0.62 & 0.00 & 0.05 & \cellcolor[HTML]{B7E4B7} 0.21 & 0.01 & 0.05 & 0.42 & \multirow{3}{*}{0.42} & \cellcolor[HTML]{B7E4B7} 1.00 & 0.11 & 0.00 & 0.06 & 0.06 & 0.04 & \cellcolor[HTML]{B7E4B7} 0.82 & 0.00 & 0.05 & 0.04 & 0.00 & 0.01 & 0.48 & \multirow{3}{*}{0.31} \\
        & $L^{-}$  & 0.11 & \cellcolor[HTML]{B7E4B7} 1.00 & 0.11 & 0.09 & 0.07 & 0.05 & 0.00 & \cellcolor[HTML]{B7E4B7} 1.00 & 0.06 & 0.00 & \cellcolor[HTML]{B7E4B7} 0.19 & 0.04 & 0.57 &  & 0.00 & \cellcolor[HTML]{B7E4B7} 0.55 & 0.06 & 0.08 & \cellcolor[HTML]{B7E4B7} 0.16 & 0.04 & 0.00 & 0.00 & 0.00 & 0.01 & \cellcolor[HTML]{B7E4B7} 0.24 & 0.04 & 0.23 &  \\
        & $L^{*}$  & 0.00 & 0.11 & \cellcolor[HTML]{B7E4B7} 1.00 & 0.06 & 0.05 & 0.08 & 0.00 & 0.00 & 0.00 & 0.05 & 0.04 & 0.05 & 0.28 &  & 0.00 & 0.10 & \cellcolor[HTML]{B7E4B7} 0.77 & 0.03 & 0.03 & 0.09 & 0.00 & 0.00 & 0.04 & 0.00 & 0.00 & 0.01 & 0.23 &  \\
        \hline

        \multirow{3}{*}{grok-3-fast}& $L^{+}$  & \cellcolor[HTML]{B7E4B7} 1.00 & 0.11 & 0.00 & \cellcolor[HTML]{B7E4B7} 0.75 & 0.19 & 0.03 & \cellcolor[HTML]{B7E4B7} 1.00 & 0.00 & 0.06 & \cellcolor[HTML]{B7E4B7} 0.68 & 0.18 & 0.05 & \textcolor{darkgreen}{\textbf{0.86}} & \multirow{3}{*}{0.53} & \cellcolor[HTML]{B7E4B7} 1.00 & 0.11 & 0.00 & \cellcolor[HTML]{B7E4B7} 0.75 & 0.19 & 0.03 & \cellcolor[HTML]{B7E4B7} 1.00 & 0.00 & 0.06 & \cellcolor[HTML]{B7E4B7} 0.68 & 0.18 & 0.05 & 0.86 & \multirow{3}{*}{0.49} \\
        & $L^{-}$  & \cellcolor[HTML]{F4CCCC} 1.00 & 0.11 & 0.00 & \cellcolor[HTML]{F4CCCC} 0.50 & 0.12 & 0.06 & 0.20 & \cellcolor[HTML]{B7E4B7} 0.29 & 0.06 & \cellcolor[HTML]{F4CCCC} 0.75 & 0.26 & 0.01 & 0.19 &  & \cellcolor[HTML]{F4CCCC} 0.83 & 0.10 & 0.05 & \cellcolor[HTML]{F4CCCC} 0.50 & 0.12 & 0.06 & 0.20 & \cellcolor[HTML]{B7E4B7} 0.29 & 0.06 & \cellcolor[HTML]{F4CCCC} 0.15 & 0.07 & 0.03 & 0.14 &  \\
        & $L^{*}$  & 0.00 & 0.11 & \cellcolor[HTML]{B7E4B7} 1.00 & 0.04 & 0.09 & \cellcolor[HTML]{B7E4B7} 0.40 & 0.04 & 0.00 & \cellcolor[HTML]{B7E4B7} 0.41 & 0.05 & 0.03 & \cellcolor[HTML]{B7E4B7} 0.40 & 0.55 &  & 0.05 & 0.05 & \cellcolor[HTML]{B7E4B7} 0.64 & 0.05 & 0.10 & \cellcolor[HTML]{B7E4B7} 0.42 & 0.00 & 0.04 & \cellcolor[HTML]{B7E4B7} 0.41 & 0.05 & 0.03 & \cellcolor[HTML]{B7E4B7} 0.42 & 0.47 &  \\
        \hline

        \multirow{3}{*}{grok-3-mini-fast}& $L^{+}$  & \cellcolor[HTML]{B7E4B7} 1.00 & 0.11 & 0.00 & \cellcolor[HTML]{B7E4B7} 0.75 & 0.19 & 0.03 & \cellcolor[HTML]{B7E4B7} 1.00 & 0.00 & 0.06 & \cellcolor[HTML]{B7E4B7} 0.68 & 0.18 & 0.05 & \textcolor{darkgreen}{\textbf{0.86}} & \multirow{3}{*}{\textcolor{darkgreen}{\textbf{0.70}}} & \cellcolor[HTML]{B7E4B7} 0.47 & 0.09 & 0.04 & \cellcolor[HTML]{B7E4B7} 0.75 & 0.19 & 0.03 & \cellcolor[HTML]{B7E4B7} 1.00 & 0.00 & 0.06 & \cellcolor[HTML]{B7E4B7} 0.68 & 0.18 & 0.05 & 0.72 & \multirow{3}{*}{0.71} \\
        & $L^{-}$  & 0.11 & \cellcolor[HTML]{B7E4B7} 1.00 & 0.11 & \cellcolor[HTML]{F4CCCC} 0.75 & 0.19 & 0.05 & 0.00 & \cellcolor[HTML]{B7E4B7} 1.00 & 0.06 & 0.15 & \cellcolor[HTML]{B7E4B7} 0.51 & 0.01 & \textcolor{darkgreen}{\textbf{0.67}} &  & 0.06 & \cellcolor[HTML]{B7E4B7} 0.42 & 0.10 & 0.19 & \cellcolor[HTML]{B7E4B7} 0.75 & 0.05 & 0.00 & \cellcolor[HTML]{B7E4B7} 1.00 & 0.06 & 0.15 & \cellcolor[HTML]{B7E4B7} 0.51 & 0.01 & 0.67 &  \\
        & $L^{*}$  & 0.04 & 0.14 & \cellcolor[HTML]{B7E4B7} 0.67 & 0.08 & 0.08 & \cellcolor[HTML]{B7E4B7} 0.45 & 0.06 & 0.06 & \cellcolor[HTML]{B7E4B7} 1.00 & 0.11 & 0.07 & \cellcolor[HTML]{B7E4B7} 0.20 & 0.58 &  & 0.00 & 0.11 & \cellcolor[HTML]{B7E4B7} 1.00 & 0.03 & 0.03 & \cellcolor[HTML]{B7E4B7} 0.63 & 0.06 & 0.06 & \cellcolor[HTML]{B7E4B7} 1.00 & 0.06 & 0.04 & \cellcolor[HTML]{B7E4B7} 0.31 & 0.74 &  \\
        \hline

        \multirow{3}{*}{kimi-k2}& $L^{+}$  & \cellcolor[HTML]{B7E4B7} 1.00 & 0.11 & 0.00 & \cellcolor[HTML]{B7E4B7} 0.62 & 0.15 & 0.02 & \cellcolor[HTML]{B7E4B7} 0.90 & 0.00 & 0.06 & \cellcolor[HTML]{B7E4B7} 0.65 & 0.18 & 0.05 & 0.79 & \multirow{3}{*}{0.52} & \cellcolor[HTML]{B7E4B7} 1.00 & 0.11 & 0.00 & \cellcolor[HTML]{B7E4B7} 0.36 & 0.08 & 0.02 & \cellcolor[HTML]{B7E4B7} 1.00 & 0.00 & 0.06 & \cellcolor[HTML]{B7E4B7} 0.13 & 0.05 & 0.04 & 0.62 & \multirow{3}{*}{0.46} \\
        & $L^{-}$  & \cellcolor[HTML]{F4CCCC} 1.00 & 0.11 & 0.00 & \cellcolor[HTML]{F4CCCC} 0.62 & 0.15 & 0.02 & 0.00 & 0.00 & 0.00 & \cellcolor[HTML]{F4CCCC} 0.48 & 0.15 & 0.04 & 0.10 &  & 0.03 & 0.03 & 0.03 & \cellcolor[HTML]{F4CCCC} 0.20 & 0.07 & 0.02 & 0.20 & \cellcolor[HTML]{B7E4B7} 0.29 & 0.06 & \cellcolor[HTML]{F4CCCC} 0.33 & 0.17 & 0.01 & 0.14 &  \\
        & $L^{*}$  & 0.00 & 0.11 & \cellcolor[HTML]{B7E4B7} 1.00 & 0.10 & 0.10 & \cellcolor[HTML]{B7E4B7} 0.36 & 0.06 & 0.06 & \cellcolor[HTML]{B7E4B7} 1.00 & 0.06 & 0.04 & \cellcolor[HTML]{B7E4B7} 0.28 & 0.66 &  & 0.00 & 0.11 & \cellcolor[HTML]{B7E4B7} 1.00 & 0.11 & 0.07 & \cellcolor[HTML]{B7E4B7} 0.43 & 0.06 & 0.06 & \cellcolor[HTML]{B7E4B7} 1.00 & 0.04 & 0.03 & 0.09 & 0.63 &  \\
        \hline

        \multirow{3}{*}{o3}& $L^{+}$  & \cellcolor[HTML]{B7E4B7} 1.00 & 0.11 & 0.00 & \cellcolor[HTML]{B7E4B7} 0.75 & 0.19 & 0.03 & \cellcolor[HTML]{B7E4B7} 1.00 & 0.00 & 0.06 & \cellcolor[HTML]{B7E4B7} 0.68 & 0.18 & 0.05 & \textcolor{darkgreen}{\textbf{0.86}} & \multirow{3}{*}{0.65} & \cellcolor[HTML]{B7E4B7} 1.00 & 0.11 & 0.00 & \cellcolor[HTML]{B7E4B7} 0.83 & 0.18 & 0.03 & \cellcolor[HTML]{B7E4B7} 1.00 & 0.00 & 0.06 & \cellcolor[HTML]{B7E4B7} 0.68 & 0.18 & 0.05 & \textcolor{darkgreen}{\textbf{0.88}} & \multirow{3}{*}{\textcolor{darkgreen}{\textbf{0.82}}} \\
        & $L^{-}$  & \cellcolor[HTML]{F4CCCC} 1.00 & 0.11 & 0.00 & \cellcolor[HTML]{F4CCCC} 0.91 & 0.22 & 0.05 & 0.00 & \cellcolor[HTML]{B7E4B7} 1.00 & 0.06 & \cellcolor[HTML]{F4CCCC} 0.52 & 0.13 & 0.02 & 0.37 &  & 0.10 & \cellcolor[HTML]{B7E4B7} 0.83 & 0.10 & 0.22 & \cellcolor[HTML]{B7E4B7} 0.91 & 0.05 & 0.00 & \cellcolor[HTML]{B7E4B7} 1.00 & 0.06 & 0.10 & \cellcolor[HTML]{B7E4B7} 0.41 & 0.01 & \textcolor{darkgreen}{\textbf{0.79}} &  \\
        & $L^{*}$  & 0.00 & 0.11 & \cellcolor[HTML]{B7E4B7} 1.00 & 0.04 & 0.06 & \cellcolor[HTML]{B7E4B7} 0.37 & 0.06 & 0.06 & \cellcolor[HTML]{B7E4B7} 1.00 & 0.03 & 0.01 & \cellcolor[HTML]{B7E4B7} 0.49 & 0.71 &  & 0.00 & 0.11 & \cellcolor[HTML]{B7E4B7} 1.00 & 0.03 & 0.05 & \cellcolor[HTML]{B7E4B7} 0.59 & 0.06 & 0.06 & \cellcolor[HTML]{B7E4B7} 1.00 & 0.02 & 0.01 & \cellcolor[HTML]{B7E4B7} 0.52 & \textcolor{darkgreen}{\textbf{0.78}} &  \\
        \hline

        \multirow{3}{*}{o4-mini}& $L^{+}$  & \cellcolor[HTML]{B7E4B7} 1.00 & 0.11 & 0.00 & \cellcolor[HTML]{B7E4B7} 0.75 & 0.19 & 0.03 & \cellcolor[HTML]{B7E4B7} 0.70 & 0.00 & 0.06 & \cellcolor[HTML]{B7E4B7} 0.68 & 0.18 & 0.05 & 0.78 & \multirow{3}{*}{0.69} & \cellcolor[HTML]{B7E4B7} 1.00 & 0.11 & 0.00 & \cellcolor[HTML]{B7E4B7} 0.75 & 0.19 & 0.03 & \cellcolor[HTML]{B7E4B7} 1.00 & 0.00 & 0.06 & \cellcolor[HTML]{B7E4B7} 0.68 & 0.18 & 0.05 & 0.86 & \multirow{3}{*}{0.78} \\
        & $L^{-}$  & 0.11 & \cellcolor[HTML]{B7E4B7} 1.00 & 0.11 & \cellcolor[HTML]{F4CCCC} 0.50 & 0.12 & 0.06 & 0.00 & \cellcolor[HTML]{B7E4B7} 1.00 & 0.06 & 0.07 & \cellcolor[HTML]{B7E4B7} 0.38 & 0.01 & 0.63 &  & 0.10 & \cellcolor[HTML]{B7E4B7} 0.83 & 0.10 & 0.15 & \cellcolor[HTML]{B7E4B7} 0.62 & 0.05 & 0.00 & \cellcolor[HTML]{B7E4B7} 1.00 & 0.06 & 0.15 & \cellcolor[HTML]{B7E4B7} 0.51 & 0.01 & 0.74 &  \\
        & $L^{*}$  & 0.00 & 0.11 & \cellcolor[HTML]{B7E4B7} 1.00 & 0.00 & 0.05 & \cellcolor[HTML]{B7E4B7} 0.42 & 0.06 & 0.06 & \cellcolor[HTML]{B7E4B7} 1.00 & 0.04 & 0.03 & \cellcolor[HTML]{B7E4B7} 0.28 & 0.68 &  & 0.00 & 0.11 & \cellcolor[HTML]{B7E4B7} 1.00 & 0.03 & 0.03 & \cellcolor[HTML]{B7E4B7} 0.71 & 0.06 & 0.06 & \cellcolor[HTML]{B7E4B7} 1.00 & 0.08 & 0.06 & \cellcolor[HTML]{B7E4B7} 0.31 & 0.75 &  \\
        \hline

        \multirow{3}{*}{qwen3-235b-a22b}& $L^{+}$  & \cellcolor[HTML]{B7E4B7} 1.00 & 0.11 & 0.00 & \cellcolor[HTML]{B7E4B7} 0.75 & 0.19 & 0.03 & \cellcolor[HTML]{B7E4B7} 1.00 & 0.00 & 0.06 & \cellcolor[HTML]{B7E4B7} 0.65 & 0.18 & 0.05 & 0.85 & \multirow{3}{*}{0.50} & \cellcolor[HTML]{B7E4B7} 1.00 & 0.11 & 0.00 & \cellcolor[HTML]{B7E4B7} 0.71 & 0.18 & 0.03 & \cellcolor[HTML]{B7E4B7} 1.00 & 0.00 & 0.06 & \cellcolor[HTML]{B7E4B7} 0.68 & 0.18 & 0.05 & 0.85 & \multirow{3}{*}{0.72} \\
        & $L^{-}$  & 0.04 & \cellcolor[HTML]{B7E4B7} 0.16 & 0.04 & \cellcolor[HTML]{F4CCCC} 0.58 & 0.16 & 0.02 & \cellcolor[HTML]{F4CCCC} 1.00 & 0.00 & 0.06 & 0.23 & \cellcolor[HTML]{B7E4B7} 0.26 & 0.01 & 0.15 &  & 0.10 & \cellcolor[HTML]{B7E4B7} 0.83 & 0.10 & 0.19 & \cellcolor[HTML]{B7E4B7} 0.60 & 0.09 & 0.00 & \cellcolor[HTML]{B7E4B7} 1.00 & 0.06 & 0.15 & \cellcolor[HTML]{B7E4B7} 0.51 & 0.01 & 0.73 &  \\
        & $L^{*}$  & 0.00 & 0.07 & \cellcolor[HTML]{B7E4B7} 0.45 & 0.09 & 0.11 & \cellcolor[HTML]{B7E4B7} 0.14 & 0.06 & 0.06 & \cellcolor[HTML]{B7E4B7} 1.00 & 0.06 & 0.02 & \cellcolor[HTML]{B7E4B7} 0.40 & 0.50 &  & 0.00 & 0.07 & \cellcolor[HTML]{B7E4B7} 0.45 & 0.04 & 0.07 & \cellcolor[HTML]{B7E4B7} 0.46 & 0.06 & 0.06 & \cellcolor[HTML]{B7E4B7} 1.00 & 0.03 & 0.02 & \cellcolor[HTML]{B7E4B7} 0.36 & 0.57 &  \\
        \hline

    \end{tabular}
    }
\end{table}

To summarize overall performance, we report average scores in the columns $\overline{diag}$ in \autoref{tab:combined_text} and \autoref{tab:combined_log}. The first subcolumn in $\overline{diag}$ shows average scores for matching pairs: ($M^+$ vs. $D^+/L^+$), ($M^-$ vs. $D^-/L^-$), and ($M^*$ vs. $D^*/L^*$). The second subcolumn in $\overline{diag}$ reports the mean of these three scores, offering a single, aggregated measure of adherence quality for each LLM. In \autoref{tab:gap_avg_diagonal}, we additionally report the score differences between the best performing LLM and each LLM’s average diagonal score.

\begin{table}[!t]
\caption{Gap to best average diagonal score ($\overline{diag}$) for each configuration. Values show the difference between the best performing LLM and each LLM's average diagonal score.}
\label{tab:gap_avg_diagonal}
    \centering
    \tiny
    \resizebox{0.9\textwidth}{!}{
    \begin{tabular}{|l|r|r|r|r|}
        \hline
        \multirow{2}{*}{LLM} & 
    \multicolumn{2}{c|}{Textual Description} & \multicolumn{2}{c|}{Log Abstraction}\\

        & Standard Prompt & Strict Prompt & Standard Prompt & Strict Prompt \\
        \hline
        command-r & 0.37 & 0.46 & 0.44 & 0.52 \\
        gemini-2.5-flash & 0.11 & 0.18 & 0.13 & 0.12 \\
        gemini-2.5-pro & 0.14 & 0.16 & 0.11 & 0.22 \\
        gpt-4.1-nano & 0.32 & 0.41 & 0.28 & 0.50 \\
        grok-3-fast & 0.05 & 0.08 & 0.17 & 0.32 \\
        grok-3-mini-fast & 0.14 & 0.05 & \cellcolor[HTML]{B7E4B7} 0.00 & 0.11 \\
        kimi-k2 & 0.16 & 0.19 & 0.19 & 0.35 \\
        o3 & 0.13 & 0.02 & 0.06 & \cellcolor[HTML]{B7E4B7} 0.00 \\
        o4-mini & \cellcolor[HTML]{B7E4B7} 0.00 & \cellcolor[HTML]{B7E4B7} 0.00 &  0.01 & 0.03 \\
        qwen3-235b-a22b & 0.10 & 0.12 & 0.21 & 0.10 \\
        \hline
    \end{tabular}
    }
\end{table}

Our findings strongly support our central hypothesis: LLMs exhibit a significant tendency for knowledge-driven hallucination when faced with atypical process structures. This is evident from two key observations. First, we identified numerous instances where models generated from reversed or shuffled artifacts were substantially more similar to the standard ground truth model ($M^+$) than to their own source evidence. As the red cells in the tables show, this phenomenon occurred across all tested LLMs, with no model achieving full adherence to atypical evidence. Second, even in cases where the LLM correctly followed the atypical process structure (green cells for $D^-$, $D^*$, $L^-$, $L^*$), the quality of the generated model, as measured by the similarity score, was generally lower than that achieved for the standard process ($D^+, L^+$). For example, while gpt-4.1-nano with the strict prompt achieved a perfect score (1.00) for the sales order process from $L^+$, the discovered models for the atypical artifacts ($L^-$ and $L^*$) received lower scores (0.55 and 0.77, respectively). This suggests that even when LLMs do not fully hallucinate, their performance is degraded when the input contradicts their internal knowledge, as they struggle to reconcile the evidence with their pre-trained schemas. 

\subsection{The Influence of Experimental Inputs and Prompts}
\textit{Effect of Strict Prompting:}
Our experiment shows that explicitly instructing the LLM to adhere strictly to the provided input can mitigate, but not eliminate, this issue. While all models were susceptible, their responsiveness to the strict prompt varied. For instance, \textit{o3} showed a marked improvement with the strict prompt, correctly modeling several atypical processes it had previously failed on. In contrast, other models like \textit{grok-3-fast} continued to hallucinate frequently even under strict instructions. With standard prompts, we observed 27 clear cases of hallucination from textual descriptions and 20 from event logs. The strict adherence prompt reduced these numbers to 13 and 10, respectively. While this improvement confirms that prompt engineering is a helpful mitigation strategy, its inability to fully resolve the problem underscores how deeply ingrained the model's background knowledge is. 

\textit{Effect of Artifact Type:}
The type of input artifact also appears to play a role. We observed fewer hallucinations when models were generated from event logs compared to textual descriptions (a total of 30 instances for logs vs. 40 for text across both prompt types). This is logical, as the structured and unambiguous format of an event log may serve as stronger evidence for the LLM compared to the inherent ambiguity of natural language. However, the persistence of the issue in log-based generation confirms that even structured data is not immune to being overridden by the model's internal schemas.

\subsection{Analysis of LLM-Specific Characteristics}
\textit{Impact of Model Properties:} Our analysis reveals that knowledge-driven hallucination is a general weakness across different LLMs, though its severity varies. Interestingly, we found no direct relationship between an LLM's size (parameter count) and its ability to adhere to atypical evidence. For instance, the massive 2.7T-parameter \textit{grok-3-fast} and the compact 18B-parameter \textit{gpt-4.1-nano} showed comparable weaknesses, while the mid-sized \textit{o4-mini} was a top performer. Similarly, while reasoning capabilities are often associated with better performance, they do not guarantee immunity to this type of hallucination. Both \textit{gemini-2.5-pro} and \textit{o4-mini} are considered reasoning models, yet \textit{o4-mini} demonstrated significantly better adherence to atypical evidence. This suggests that neither raw scale nor general reasoning ability alone predicts a model's fidelity to source evidence when it conflicts with pre-trained knowledge.

\textit{Correlation Between General Capability and Knowledge Hallucination:}
A particularly revealing finding is that high performance on standard tasks does not guarantee robustness against knowledge hallucination. A prime example is \textit{gemini-2.5-pro}, which shows strong performance on a wide range of tasks (as measured by the LiveBench benchmark). It was also the top-performing model in our experiment when using the standard prompt and the textual descriptions ($D^+$), achieving the highest average score (0.80). However, its performance collapsed when faced with conflicting artifacts, dropping to 0.33 for the reversed descriptions ($D^-$) and just 0.11 for the shuffled ones ($D^*$). This dramatic drop suggests that the model's well-formed internal schema for the standard process is so dominant that it consistently overrides conflicting source evidence, making it highly prone to knowledge-driven hallucination.

\section{Conclusion}
\label{sec:conclusion}

This paper introduced and empirically investigated the phenomenon of \emph{knowledge-driven hallucination} in Large Language Models (LLMs), where a model's pre-trained knowledge overrides explicit source evidence, leading to factually incorrect but plausible-looking outputs. Through a controlled experiment in the domain of automated process modeling, we systematically evaluated the fidelity of ten state-of-the-art LLMs when tasked with generating process models from standard and deliberately atypical process evidence.

Our findings demonstrate that LLMs exhibit a tendency to prioritize their generalized internal schemas over contradictory evidence provided in the prompt. This was evident as models frequently reverted to generating a standard process flow even when the input described a reversed or structurally shuffled version. We observed this behavior across all tested LLMs, regardless of their size or specialization, and with both unstructured text and structured event log inputs. Even in cases where the models did not fully hallucinate, their performance in correctly modeling atypical processes was significantly degraded compared to standard ones, highlighting the disruptive effect of the conflict between evidence and internal knowledge.

The implications of our findings extend far beyond process modeling and raise critical concerns about the reliability of LLMs in any evidence-based domain. The danger of knowledge-driven hallucination lies in its deceptive nature; the generated artifacts are often coherent, logical, and well-formed, masking the fact that they do not accurately represent the source data. This ``plausibility trap'' poses a significant risk in fields such as legal analysis, financial reporting, and scientific research, where strict adherence to source evidence is essential. Our work underscores that simple prompt engineering, such as instructing the model to be faithful to the input, can mitigate but not eliminate this deep-seated behavior.

Future work should focus on two key areas. First, there is a clear need to develop more robust mitigation techniques beyond prompting that allow for better control over the influence of pre-trained knowledge. Second, this experimental methodology could be adapted to investigate knowledge-driven hallucination in other structured generation tasks, such as code generation from legacy specifications or data schema creation from business requirements. Ultimately, our study serves as a critical reminder that as we delegate more complex analytical tasks to AI, we must also develop rigorous methods to validate its outputs and ensure that its powerful inferential capabilities do not come at the cost of factual integrity.

\textbf{Acknowledgments.} 
This work was funded by the Federal Ministry of Research, Technology and Space (BMFTR), Germany (grant 01IS23065).

\bibliographystyle{splncs04}
\bibliography{shortlit}

@inproceedings{DBLP:conf/nips/BrownMRSKDNSSAA20,
  author       = {Tom B. Brown and
                  Benjamin Mann and
                  Nick Ryder and
                  Melanie Subbiah et al.},
  bibeditor       = {Hugo Larochelle and
                  Marc'Aurelio Ranzato and
                  Raia Hadsell and
                  Maria{-}Florina Balcan and
                  Hsuan{-}Tien Lin},
  title        = {Language Models are Few-Shot Learners},
  booktitle    = {NeurIPS 2020},
  year         = {2020},
  _url          = {https://proceedings.neurips.cc/paper/2020/hash/1457c0d6bfcb4967418bfb8ac142f64a-Abstract.html},
  timestamp    = {Thu, 25 May 2023 10:38:31 +0200},
  biburl       = {https://dblp.org/rec/conf/nips/BrownMRSKDNSSAA20.bib},
  bibsource    = {dblp computer science bibliography, https://dblp.org}
}

@inproceedings{DBLP:conf/bpm/GrohsAER23,
  author       = {Michael Grohs and
                  Luka Abb and
                  Nourhan Elsayed and
                  Jana{-}Rebecca Rehse},
  bibeditor       = {Jochen De Weerdt and
                  Luise Pufahl},
  title        = {Large Language Models Can Accomplish Business Process Management Tasks},
  booktitle    = {{BPM} 2023 Workshops},
  series       = {LNBIP },
  volume       = {492},
  pages        = {453--465},
  publisher    = {Springer},
  year         = {2023},
  _url          = {https://doi.org/10.1007/978-3-031-50974-2\_34},
  timestamp    = {Wed, 31 Jan 2024 16:29:47 +0100},
  biburl       = {https://dblp.org/rec/conf/bpm/GrohsAER23.bib},
  bibsource    = {dblp computer science bibliography, https://dblp.org}
}

@inproceedings{DBLP:conf/bpmds/KouraniB0A24,
  author       = {Humam Kourani and
                  Alessandro Berti and
                  Daniel Schuster and
                  Wil M. P. van der Aalst},
  title        = {Process Modeling with Large Language Models},
  booktitle    = {Enterprise, Business-Process and Information Systems Modeling - {BPMDS} 2024 and {EMMSAD} 2024, Limassol, Cyprus, June 3-4, 2024, Proceedings},
  pages        = {229--244},
  year         = {2024},
  bibcrossref     = {DBLP:conf/bpmds/2024},
  bibdoi          = {10.1007/978-3-031-61007-3\_18},
  bibsource    = {dblp computer science bibliography, https://dblp.org}
}

@inproceedings{DBLP:conf/ijcai/KouraniB0A24,
  author       = {Humam Kourani and
                  Alessandro Berti and
                  Daniel Schuster and
                  Wil M. P. van der Aalst},
  title        = {{ProMoAI}: Process Modeling with Generative {AI}},
  booktitle    = {{IJCAI} 2024, Jeju, South Korea, August 3-9, 2024},
  pages        = {8708--8712},
  year         = {2024},
  bibcrossref     = {DBLP:conf/ijcai/2024},
  bibsource    = {dblp computer science bibliography, https://dblp.org}
}

@article{DBLP:journals/jucs/GoncalvesSB11,
  author       = {Jo{\~{a}}o Carlos de A. R. Gon{\c{c}}alves and
                  Fl{\'{a}}via Maria Santoro and
                  Fernanda Ara{\'{u}}jo Bai{\~{a}}o},
  title        = {Let Me Tell You a Story - On How to Build Process Models},
  journal      = {J. Univers. Comput. Sci.},
  volume       = {17},
  number       = {2},
  pages        = {276--295},
  year         = {2011},
  _url          = {https://doi.org/10.3217/jucs-017-02-0276},
  timestamp    = {Thu, 15 Feb 2024 13:27:26 +0100},
  biburl       = {https://dblp.org/rec/journals/jucs/GoncalvesSB11.bib},
  bibsource    = {dblp computer science bibliography, https://dblp.org}
}

@inproceedings{DBLP:conf/caise/0003W0LLZLW20,
  author       = {{Chen Qian et al.}},
  bibeditor       = {Schahram Dustdar and
                  Eric Yu and
                  Camille Salinesi and
                  Dominique Rieu and
                  Vik Pant},
  title        = {An Approach for Process Model Extraction by Multi-grained Text Classification},
  booktitle    = {CAiSE 2020},
  series       = {LNCS},
  volume       = {12127},
  pages        = {268--282},
  publisher    = {Springer},
  year         = {2020},
  _url          = {https://doi.org/10.1007/978-3-030-49435-3\_17},
  timestamp    = {Thu, 20 May 2021 08:59:37 +0200},
  biburl       = {https://dblp.org/rec/conf/caise/0003W0LLZLW20.bib},
  bibsource    = {dblp computer science bibliography, https://dblp.org}
}

@book{DBLP:books/sp/DumasRMR18,
  author       = {Marlon Dumas and
                  Marcello La Rosa and
                  Jan Mendling and
                  Hajo A. Reijers},
  title        = {Fundamentals of Business Process Management, Second Edition},
  publisher    = {Springer},
  year         = {2018},
  _url          = {https://doi.org/10.1007/978-3-662-56509-4},
  isbn         = {978-3-662-56508-7},
  timestamp    = {Wed, 07 Dec 2022 23:14:28 +0100},
  biburl       = {https://dblp.org/rec/books/sp/DumasRMR18.bib},
  bibsource    = {dblp computer science bibliography, https://dblp.org}
}

@inproceedings{DBLP:conf/bpm/KouraniZ23,
  author       = {Humam Kourani and
                  Sebastiaan J. van Zelst},
  title        = {{POWL:} Partially Ordered Workflow Language},
  booktitle    = {{BPM}
                  2023},
  pages        = {92--108},
  year         = {2023},
  bibcrossref     = {DBLP:conf/bpm/2023},
  bibdoi          = {10.1007/978-3-031-41620-0\_6},
  bibsource    = {dblp computer science bibliography, https://dblp.org}
}

@article{DBLP:journals/simpa/BertiZS23,
  author       = {Alessandro Berti and
                  Sebastiaan J. van Zelst and
                  Daniel Schuster},
  title        = {{PM4Py}: {A} process mining library for Python},
  journal      = {Softw. Impacts},
  volume       = {17},
  pages        = {100556},
  year         = {2023},
  bibdoi          = {10.1016/J.SIMPA.2023.100556},
  bibsource    = {dblp computer science bibliography, https://dblp.org}
}

@inproceedings{DBLP:conf/caise/FriedrichMP11,
  author       = {Fabian Friedrich and
                  Jan Mendling and
                  Frank Puhlmann},
  title        = {Process Model Generation from Natural Language Text},
  booktitle    = {CAiSE 2011},
  pages        = {482--496},
  year         = {2011},
  bibcrossref     = {DBLP:conf/caise/2011},
  bibdoi          = {10.1007/978-3-642-21640-4\_36},
  bibsource    = {dblp computer science bibliography, https://dblp.org}
}

@article{DBLP:journals/eis/WoenselM24,
  author       = {William Van Woensel and
                  Soroor Motie},
  title        = {{NLP4PBM:} a systematic review on process extraction using natural
                  language processing with rule-based, machine and deep learning methods},
  journal      = {Enterp. Inf. Syst.},
  volume       = {18},
  number       = {11},
  year         = {2024},
  bibdoi          = {10.1080/17517575.2024.2417404},
  bibsource    = {dblp computer science bibliography, https://dblp.org}
}

@article{DBLP:journals/jksucis/SholiqSA22,
  author       = {Sholiq Sholiq and
                  Riyanarto Sarno and
                  Endang Siti Astuti},
  title        = {Generating {BPMN} diagram from textual requirements},
  journal      = {J. King Saud Univ. Comput. Inf. Sci.},
  volume       = {34},
  number       = {10 Part {B}},
  pages        = {10079--10093},
  year         = {2022},
  bibdoi          = {10.1016/J.JKSUCI.2022.10.007},
  bibsource    = {dblp computer science bibliography, https://dblp.org}
}

@inproceedings{DBLP:conf/wecwis/SintorisV17,
  author       = {Konstantinos Sintoris and
                  Kostas Vergidis},
  title        = {Extracting Business Process Models Using Natural Language Processing
                  {(NLP)} Techniques},
  booktitle    = {{CBI} 2017},
  pages        = {135--139},
  year         = {2017},
  bibcrossref     = {DBLP:conf/wecwis/2017-1},
  bibdoi          = {10.1109/CBI.2017.41},
  bibsource    = {dblp computer science bibliography, https://dblp.org}
}

@article{DBLP:journals/csur/JiLFYSXIBMF23,
  author       = {{Ziwei Ji et al.}},
  title        = {Survey of Hallucination in Natural Language Generation},
  journal      = {{ACM} Comput. Surv.},
  volume       = {55},
  number       = {12},
  pages        = {248:1--248:38},
  year         = {2023},
  bibdoi          = {10.1145/3571730},
  bibsource    = {dblp computer science bibliography, https://dblp.org}
}

@article{DBLP:journals/corr/abs-2410-03255,
  author       = {Kiran Busch and
                  Henrik Leopold},
  title        = {Towards a Benchmark for Large Language Models for Business Process
                  Management Tasks},
  journal      = {CoRR},
  volume       = {abs/2410.03255},
  year         = {2024},
  bibdoi          = {10.48550/ARXIV.2410.03255},
  eprinttype    = {arXiv},
  eprint       = {2410.03255},
  bibsource    = {dblp computer science bibliography, https://dblp.org}
}

@inproceedings{DBLP:conf/bpm/KlievtsovaBKMR23,
  author       = {Nataliia Klievtsova and
                  Janik{-}Vasily Benzin and
                  Timotheus Kampik and
                  Juergen Mangler and
                  Stefanie Rinderle{-}Ma},
  title        = {Conversational Process Modelling: State of the Art, Applications,
                  and Implications in Practice},
  booktitle    = {{BPM} 2023 Forum},
  pages        = {319--336},
  year         = {2023},
  bibcrossref     = {DBLP:conf/bpm/2023f},
  bibdoi          = {10.1007/978-3-031-41623-1\_19},
  bibsource    = {dblp computer science bibliography, https://dblp.org}
}

@inproceedings{DBLP:conf/caise/ForsterPW13,
  author       = {Simon Forster and
                  Jakob Pinggera and
                  Barbara Weber},
  title        = {Toward an Understanding of the Collaborative Process of Process Modeling},
  booktitle    = {CAiSE'13 Forum},
  pages        = {98--105},
  year         = {2013},
  bibcrossref     = {DBLP:conf/caise/2013fo},
  bibsource    = {dblp computer science bibliography, https://dblp.org}
}

@inproceedings{DBLP:conf/coling/AaCLMP18,
  author       = {Han van der Aa and
                  Josep Carmona and
                  Henrik Leopold and
                  Jan Mendling and
                  Llu{\'{\i}}s Padr{\'{o}}},
  title        = {Challenges and Opportunities of Applying Natural Language Processing
                  in Business Process Management},
  booktitle    = {{COLING} 2018},
  pages        = {2791--2801},
  year         = {2018},
  bibcrossref     = {DBLP:conf/coling/2018},
  bibsource    = {dblp computer science bibliography, https://dblp.org}
}

@article{DBLP:journals/corr/abs-2412-00023,
  author       = {Humam Kourani and
                  Alessandro Berti and
                  Daniel Schuster and
                  Wil M. P. van der Aalst},
  title        = {Evaluating Large Language Models on Business Process Modeling: Framework,
                  Benchmark, and Self-Improvement Analysis},
  journal      = {CoRR},
  volume       = {abs/2412.00023},
  year         = {2024},
  bibdoi          = {10.48550/ARXIV.2412.00023},
  eprinttype    = {arXiv},
  eprint       = {2412.00023},
  bibsource    = {dblp computer science bibliography, https://dblp.org}
}

@article{DBLP:journals/tois/HuangYMZFWCPFQL25,
  author       = {{Lei Huang et al.}},
  title        = {A Survey on Hallucination in Large Language Models: Principles, Taxonomy,
                  Challenges, and Open Questions},
  journal      = {{ACM} Trans. Inf. Syst.},
  volume       = {43},
  number       = {2},
  pages        = {42:1--42:55},
  year         = {2025},
  bibdoi          = {10.1145/3703155},
  bibsource    = {dblp computer science bibliography, https://dblp.org}
}

@inproceedings{DBLP:conf/nips/ZhouLX0SMMEYYZG23,
  author       = {{Chunting Zhou et al.}},
  bibeditor       = {Alice Oh and
                  Tristan Naumann and
                  Amir Globerson and
                  Kate Saenko and
                  Moritz Hardt and
                  Sergey Levine},
  title        = {{LIMA:} Less Is More for Alignment},
  booktitle    = {NeurIPS 2023},
  year         = {2023},
  bibsource    = {dblp computer science bibliography, https://dblp.org}
}

@article{DBLP:journals/corr/abs-2307-12966,
  author       = {{Yufei Wang et al.}},
  title        = {Aligning Large Language Models with Human: {A} Survey},
  journal      = {CoRR},
  volume       = {abs/2307.12966},
  year         = {2023},
  bibdoi          = {10.48550/ARXIV.2307.12966},
  eprinttype    = {arXiv},
  eprint       = {2307.12966},
  bibsource    = {dblp computer science bibliography, https://dblp.org}
}

@article{DBLP:journals/corr/abs-2101-00027,
  author       = {{Leo Gao et al.}},
  title        = {The Pile: An 800GB Dataset of Diverse Text for Language Modeling},
  journal      = {CoRR},
  volume       = {abs/2101.00027},
  year         = {2021},
  eprinttype    = {arXiv},
  eprint       = {2101.00027},
  bibsource    = {dblp computer science bibliography, https://dblp.org}
}

@article{DBLP:journals/corr/abs-2305-18703,
  author       = {{Chen Ling et al.}},
  title        = {Beyond One-Model-Fits-All: {A} Survey of Domain Specialization for
                  Large Language Models},
  journal      = {CoRR},
  volume       = {abs/2305.18703},
  year         = {2023},
  bibdoi          = {10.48550/ARXIV.2305.18703},
  eprinttype    = {arXiv},
  eprint       = {2305.18703},
  bibsource    = {dblp computer science bibliography, https://dblp.org}
}

@article{roberts2024extending,
  title={Extending contextual length and world knowledge generalization in large language models},
  author={Roberts, Malajah and Anderson, Jonathan and Delgado, William and Johnson, Richard and Spencer, Lawrence},
  year={2024}
}

@inproceedings{DBLP:conf/acl/ChenLHZSLLY24,
  author       = {{Longze Chen et al.}},
  bibeditor       = {Lun{-}Wei Ku and
                  Andre Martins and
                  Vivek Srikumar},
  title        = {Long Context is Not Long at All: {A} Prospector of Long-Dependency
                  Data for Large Language Models},
  booktitle    = {{ACL} 2024},
  pages        = {8222--8234},
  publisher    = {Association for Computational Linguistics},
  year         = {2024},
  bibdoi          = {10.18653/V1/2024.ACL-LONG.447},
  bibsource    = {dblp computer science bibliography, https://dblp.org}
}

@inproceedings{DBLP:conf/iclr/HoskingBB24,
  author       = {Tom Hosking and
                  Phil Blunsom and
                  Max Bartolo},
  title        = {Human Feedback is not Gold Standard},
  booktitle    = {{ICLR} 2024},
  publisher    = {OpenReview.net},
  year         = {2024},
  bibsource    = {dblp computer science bibliography, https://dblp.org}
}

@inproceedings{DBLP:conf/iclr/SharmaTKDABDHJK24,
  author       = {{Mrinank Sharma et al.}},
  title        = {Towards Understanding Sycophancy in Language Models},
  booktitle    = {{ICLR} 2024},
  publisher    = {OpenReview.net},
  year         = {2024},
  bibsource    = {dblp computer science bibliography, https://dblp.org}
}

@article{DBLP:journals/air/BlancoJusticiaJMSDCT25,
  author       = {{Alberto Blanco{-}Justicia
                  et al.}},
  title        = {Digital forgetting in large language models: a survey of unlearning
                  methods},
  journal      = {Artif. Intell. Rev.},
  volume       = {58},
  number       = {3},
  pages        = {90},
  year         = {2025},
  bibdoi          = {10.1007/S10462-024-11078-6},
  bibsource    = {dblp computer science bibliography, https://dblp.org}
}

@inproceedings{DBLP:conf/iclr/BerglundTKBSKE24,
  author       = {{Lukas Berglund et al.}},
  title        = {The Reversal Curse: LLMs trained on "A is B" fail to learn "B is A"},
  booktitle    = {{ICLR} 2024},
  publisher    = {OpenReview.net},
  year         = {2024},
  bibsource    = {dblp computer science bibliography, https://dblp.org}
}

\end{document}